\documentclass[twocolumn,floatfix,aps,superscriptaddress,prb]{revtex4-1}

\usepackage[utf8]{inputenc}
\usepackage{natbib}
\usepackage{graphicx}
\usepackage{xcolor}
\usepackage{bm}
\usepackage{mhchem}
\usepackage{physics}
\usepackage{amsfonts}
\usepackage{amssymb}
\usepackage{scalerel}
\newcommand{\smallsquare}{\scaleobj{0.7}{\square}}
\usepackage[normalem]{ulem}
\usepackage{comment}

\usepackage{pdfpages}
\usepackage{pgffor}
\makeatletter
\AtBeginDocument{\let\LS@rot\@undefined}
\makeatother

\definecolor{cocol}{rgb}{0,0.6,0}

\definecolor{bluish}{rgb}{0,0.4,0.9}

\newcommand{\rev}[1]{{#1}}

\makeatother

\ifdefined\DIRACREP
\else
\newcommand{\DIRACREP}{}

\usepackage{physics}
\usepackage{xparse}
\ifdefined\COSMOMATHS
\else
\newcommand{\COSMOMATHS}{}

\newcommand{\mbf}[1]{\ensuremath{\mathbf{#1}}}

\fi %

\NewDocumentCommand{\rep}{s d<| d|>}{%
\IfBooleanTF{#1}{
   \IfValueTF{#2}{
       \IfValueTF{#3}{\braket{#2}{#3}}{\bra{#2}}
       }{
       \IfValueTF{#3}{\ket{#3}}{}
       }
   }{
   \IfValueTF{#2}{
       \IfValueTF{#3}{\braket*{#2}{#3}}{\bra*{#2}}
       }{
       \IfValueTF{#3}{\ket*{#3}}{}
       }
   }
}

\NewDocumentCommand{\rbra}{sm}{\IfBooleanTF{#1}{\rep*<#2|}{\rep<#2|}}
\NewDocumentCommand{\rket}{sm}{\IfBooleanTF{#1}{\rep*|#2>}{\rep|#2>}}
\NewDocumentCommand{\rbraket}{smom}{
    \IfBooleanTF{#1}{
        \IfNoValueTF{#3}{\rep*<#2||#4>}{\rep*<#2|#3\rep*|#4>}
    }{
        \IfNoValueTF{#3}{\rep<#2||#4>}{\rep<#2|#3\rep|#4>}
    }
}

\NewDocumentCommand{\field}{o m e{_} e{^} o e{_} e{^}}{
\IfValueTF{#5}{\overline{
  #2\IfValueT{#3}{_#3}\IfValueT{#4}{^{\otimes #4}} %
  \otimes
  #5\IfValueT{#6}{_#6}\IfValueT{#7}{^{\otimes #7}} %
  \IfValueT{#1}{;#1}
}}{
  \IfValueTF{#4}{\overline{
     #2\IfValueT{#3}{_#3}\IfValueT{#4}{^{\otimes #4}}
     \IfValueT{#1}{;#1}
  }}
  {#2\IfValueT{#3}{_#3}}
}
}

\NewDocumentCommand{\frho}{o e{_} e{^}}{
\field[#1]{\rho}_{#2}^{#3}
}

\newcommand{\br}{\mbf{r}}
\newcommand{\bx}{\mbf{x}}

\newcommand{\e}{a}  %

\NewDocumentCommand{\ex}{e_}{
\IfValueTF{#1}{\e_{#1}\bx_{#1}}{\e\bx}
}  %

\NewDocumentCommand{\lm}{e_}{
\IfValueTF{#1}{l_{#1}m_{#1}}{lm}
}
\NewDocumentCommand{\nlm}{e_}{
\IfValueTF{#1}{n_{#1}\lm_{#1}}{n\lm}
}
\NewDocumentCommand{\enlm}{e_}{
\IfValueTF{#1}{\e_{#1}\nlm_{#1}}{\e\nlm}
}
\NewDocumentCommand{\en}{e_}{
\IfValueTF{#1}{\e_{#1}n_{#1}}{\e n}
}
\NewDocumentCommand{\nlk}{e_}{
\IfValueTF{#1}{n_{#1}l_{#1}k_{#1}}{nlk}
}
\NewDocumentCommand{\enlk}{e_}{
\IfValueTF{#1}{\e_{#1}\nlk_{#1}}{\e\nlk}
}
\NewDocumentCommand{\enl}{e_}{
\IfValueTF{#1}{\en_{#1}l_#1}{\en l}
}

\NewDocumentCommand{\nnl}{s}{
\IfBooleanTF{#1}{n_1 n_2 l}{n_1; n_2; l}
}
\NewDocumentCommand{\ennl}{s}{
\IfBooleanTF{#1}{\en_1 \en_2 l}{\en_1; \en_2; l}
}

\NewDocumentCommand{\gslm}{s}{
\IfBooleanTF{#1}{\sigma\lambda\mu}{\sigma;\lambda\mu}
}

\fi %

\ifdefined\COSMOMODELS
\else
\newcommand{\COSMOMODELS}{}
\usepackage{bm,upgreek}

\newcommand{\krn}[0]{\operatorname{k}}
\newcommand{\dst}[0]{\operatorname{d}}

\newcommand{\feat}{\upxi}
\newcommand{\bfeat}[0]{\ensuremath{\bm{\upxi}}}

\fi %

\begin{document}

\title{Incompleteness of graph neural networks\\ for points clouds in three dimensions}

\author{Sergey N. Pozdnyakov}
\affiliation{Laboratory of Computational Science and Modelling, Institute of Materials, Ecole Polytechnique F\'ed\'erale de Lausanne, Lausanne 1015, Switzerland}
\author{Michele Ceriotti}
\email{michele.ceriotti@epfl.ch}
 \affiliation{Laboratory of Computational Science and Modelling, Institute of Materials, Ecole Polytechnique F\'ed\'erale de Lausanne, Lausanne 1015, Switzerland}
\date{\today}

\begin{abstract}
Graph neural networks (GNN) are very popular methods in machine learning and have been applied very successfully to the prediction of the properties of molecules and materials. First-order GNNs are well known to be incomplete, i.e., there exist graphs that are distinct but appear identical when seen through the lens of the GNN. More complicated schemes have thus been designed to increase their resolving power. 
Applications to molecules (and more generally, point clouds), however, add a geometric dimension to the problem. The most straightforward and prevalent approach to construct graph representation for molecules regards atoms as vertices in a graph and draws a bond between each pair of atoms within a chosen cutoff. 
Bonds can be decorated with the distance between atoms, and the resulting ``distance graph NNs'' (dGNN) have empirically demonstrated excellent resolving power and are widely used in chemical ML, {with all known indistinguishable configurations being resolved in the fully-connected limit, which is equivalent to infinite or sufficiently large cutoff.}
Here we {present a counterexample that proves that dGNNs are not complete even for the restricted case of fully-connected graphs induced by 3D atom clouds.} We construct pairs of distinct point clouds whose associated graphs are, for any cutoff radius, equivalent based on a first-order Weisfeiler-Lehman test. 
This class of degenerate structures includes chemically-plausible configurations, {both for isolated structures and for infinite structures that are periodic in 1, 2, and 3 dimensions.}  The existence of indistinguishable configurations sets an ultimate limit to the expressive power of some of the well-established GNN architectures for atomistic machine learning. Models that explicitly use angular or directional information in the description of atomic environments can resolve this class of degeneracies.
\end{abstract}

\maketitle

\section{Introduction}

Point clouds can be used to provide an abstract description of shapes, and objects across different length scales, \cite{gumh+01imr,guo+21tpami} and have been widely used in the construction of machine-learning models for computer vision \cite{wu+19ccvpr}, remote sensing \cite{bell+20rs} and autonomous driving \cite{li+21tnnls}.
A description in terms of an unordered set of points is relevant also at the atomic scale, where molecules and materials are most naturally characterised in terms of the position and nature of their atomic constituents. 
The list of Cartesian coordinates of points, however, does not reflect the fact that the properties associated with a given structure are usually invariant, or equivariant, to symmetry operations such as rigid translations, rotations, or permutation of the ordering of the points.
In the context of atomistic simulations, the problem of describing a structure in a symmetry-adapted manner has been a central concern in early applications of machine-learning models  \cite{behl-parr07prl,bart+10prl,rupp+12prl}, and has since given rise to the development of a large number of \emph{representations} that attempt to characterize fully a structure while simultaneously fulfilling  the requirements of symmetry \cite{musi+21cr}.

In fact, deep connections are present between most of the existing frameworks \cite{will+19jcp}, that differ in implementation details but can be understood as describing structures in terms of unordered lists of distances, angles, tetrahedra, etc. within atom-centered environments, corresponding to correlations between 2, 3, 4, $\nu$ neighbors of the central atom. 
This systematic study has revealed that -- at least for low values of $\nu$ -- atom-centered representations are \emph{incomplete}, i.e. there are pairs of structures that are distinct, but contain environments that are indistinguishable based on the unordered list of distances and angles \cite{pozd+20prl}, affecting also the ability of the representation to resolve local deformations of certain structures \cite{pozd+21ore}. 

\begin{figure*}[bthp]
    \centering
    \includegraphics[width=0.9\linewidth]{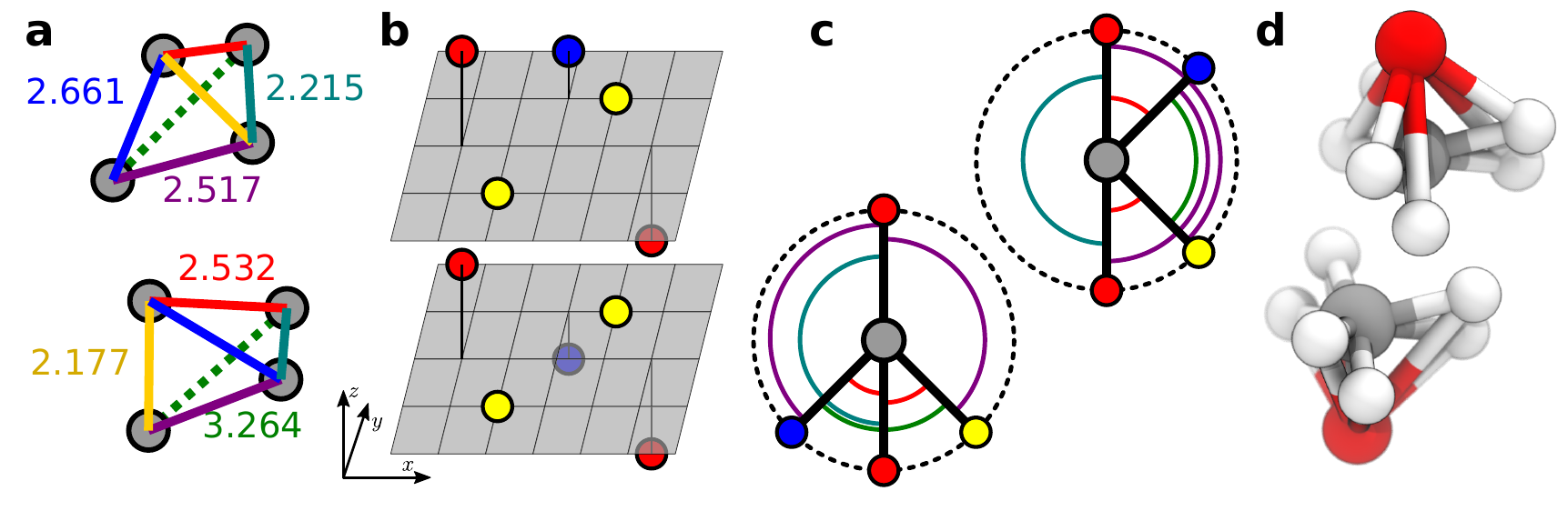}
    \caption{Examples of structures (a,b), or environments (c,d) that are distinct but cannot be discriminated by the unordered list of distances or distances and angles between the points. (a) The two tetrahedra share the same list of pair distances  $\{ r_{ij} \}_{i,j\in A}$, as per color coding.   (b) The two structure share the same list of pair distances, and in addition the list of distances of each atom relative to its neighbors $\{ \{ r_{ij} \}_{j\in A} \}_{i\in A}$. (c) The two environments share the same list of distances and angles relative to the central (gray) atom, $\{ (r_{1j}, r_{1j'}, \br_{1j}\cdot\br_{1j'})  \}_{j,j'\in A}  $. (d) The two environments share the same list of distances, angles and tetrahedra $\{ (r_{1j}, r_{1j'},r_{1j''}, \br_{1j}\cdot\br_{1j'}, \br_{1j}\cdot\br_{1j''}, \br_{1j'}\cdot\br_{1j''})  \}_{j,j',j''\in A}  $ around the central (gray) atom. The example (a) is taken from \cite{bart+13prb}, while (b,c,d) are from \cite{pozd+20prl} }
    \label{fig:classics}
\end{figure*}

\begin{figure}[bp]
\centering
   \includegraphics[width=0.9\linewidth]{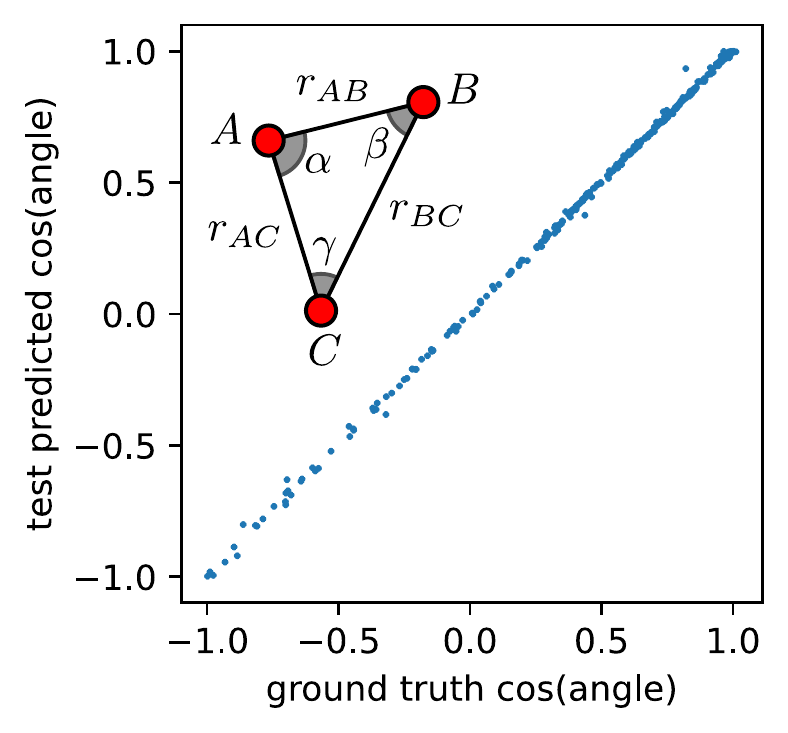} 
   \caption{{A simple demonstration of the ability of dGNN to encode information beyond interatomic distances. The parity plot demonstrates that SchNet can learn the angles associated with the vertices of triangles with variable shape, despite using only distances as inputs. \label{fig:triangle} }}
\end{figure}

This issue is also closely related to classical problems in invariant theory \cite{bout-kemp04aam}, that aim to determine under which conditions two point clouds can be unequivocally identified by the unordered list of distances, or distances and angles.
If a set of features cannot discriminate two structures, any metric\cite{sade+13jcp,widd+22cmcc} built on those features will not be able to distinguish them.
Furthermore any model built on those features - no matter how sophisticated - will be unable to learn their properties or to classify them in different categories, and will be in general terms limited in its expressive power. 
The actual impact of these degeneracies on machine-learning models of atomic and materials properties is mitigated by the fact that structures that contain problematic environments usually also contain others that are not degenerate, and so a model that \emph{simultaneously} uses information on all centers can still distinguish the structures \cite{vonl+15ijqc}. 
For instance, all structures in Fig.~\ref{fig:classics} except for (b) can be discriminated by knowledge of the set of unordered lists of distances $r_{ij}=|\br_i-\br_j|$ around each point (i.e. taking $\{ \{ r_{ij} \}_{j\in A} \}_{i\in A}$. 
Still, there is evidence that the incompleteness of atom-density representations does affect the ultimate performance of the machine-learning models built on them. \cite{pozd+20prl} This is therefore an issue of practical, and not only theoretical, relevance, which is one of the factors driving a transition towards systematically-improvable features with higher correlation order \cite{shap16mms,will+19jcp,drau19prb,niga+20jcp}. 

In the broader context of models based on point clouds, graph neural networks (GNN) have been used extensively to describe the relative arrangement of points, mapping the cloud onto a graph. \cite{wang+19acmtg,wu+19ccvpr} 
In the case of atomistic models, most well-established GNN frameworks treat atoms as the vertices of the graph, labeled by their chemical identity, and the edges are associated with the distances between them.\cite{gilm+17icml,schu+18jcp,schu+17nips}
Even though this class of ``distance graph'' NN (dGNN) only uses information on the distances between points, the way this is combined in the subsequent steps of the network allows for very flexible models with considerable descriptive power. {One simple but compelling example is given in Fig.~\ref{fig:triangle}. By applying a dGNN to a set of triangular configurations, one can predict the angles even though the network uses only distance information. As we shall see, this is not generally true: for configurations with many points it is not always possible to infer angular information using a dGNN.  
}

It is well-known that for general discrete graphs it is possible to build pairs of items that are distinct, and yet indistinguishable by most traditional GNN. \cite{sato2020arxiv} This has triggered the development of higher-order graph networks \cite{morr+18arxiv,thom+18arxiv}. 
Examples of actual chemical structures that cannot be distinguished by a dGNN can also be obtained if one considers a finite cutoff \cite{zhan+21prl,vikas2020arxiv}.  
To the best of our knowledge, however, no examples have been shown of 3D point clouds whose \emph{fully-connected}, distance-labelled graphs cannot be resolved by a dGNN.
In fact, with a sufficiently long cutoff in the construction of the graph, dGNNs can discriminate between all examples of environments that are degenerate under $\nu=1,2,3$ atom-centered descriptors shown in Fig.~\ref{fig:classics}, as well as those in  \cite{zhan+21prl,vikas2020arxiv}, suggesting a higher descriptive power than atom-centered descriptors. 
Here we present a class of point clouds involving pairs of structures that are not distinguishable by dGNN, irrespective of the cutoff chosen in the construction of the molecular graph. 
We show that this construction includes also configurations that correspond to realistic chemical structures, and demonstrate how for these configurations there is a limit to the accuracy that can be reached by this class of models.

\section{Weisfeiler-Lehman test for geometric GNNs}
\label{sec:w-l-test}

The Weisfeiler-Lehman test is a well-known graph-theoretical procedure \cite{weis-lehm68NTI,douglas2011arxiv} that provides a sufficient condition for two graphs being distinct, and that has been shown to be equivalent to an assessment of the resolving power of several classes of GNNs \cite{sher-borg09nips,kipf2016arxiv,sato2020arxiv}. 
Here we use a version of the test that incorporates explicitly the distances between nodes in the construction of the edge identifiers, and is similar in spirit to the construction of graph kernels in Ref.~\citenum{sher+11jmlr}.
Given a set of $n$ nodes with labels, $\{l_i\}_{i=1\ldots n}$, and distances between them and their neighbors $\{r_{ij}\}$, we construct fingerprints of each node as the multiset of label/distance pairs, supplemented with a label for the node
\begin{equation}
h_i= \operatorname{hash}(l_i,\{\{ (l_j, r_{ij})\}\}_{j=1\ldots n}).
\end{equation}
The hash function assigns a unique identifier to distinct multisets. The values of the hashes are then used to re-label the nodes, and the procedure is iterated: the process of encoding in the node labels the structure of the neighborhood and then iterating the procedure has a clear analogy with message passing constructs. 
{Returning to the example in Fig.~\ref{fig:triangle}, it should now be clear how, for a triangle, a dGNN can learn angular information. After two iterations, the node descriptors for each of the vertices contain information not only on the two distances to the first neighbors (e.g., for vertex $A$, $r_{AB}$ and $r_{AC}$), but also on the fact that the neighbors have a further neighbor at a certain distance (e.g. $r_{BC}$), which is all the information needed to reconstruct the angles. 
}
Usually (but not always) the iterative procedure converges to fixed values of the vertex labels; if the multisets of hashes that characterize two graphs are identical, the graphs cannot be distinguished by this distance-decorated WL test, and won't be discriminated by a geometric dGNN that uses only point labels and pair distances to characterize graph neighborhoods. 

\begin{figure}[tb]
    \centering
    \includegraphics[width=0.8\linewidth]{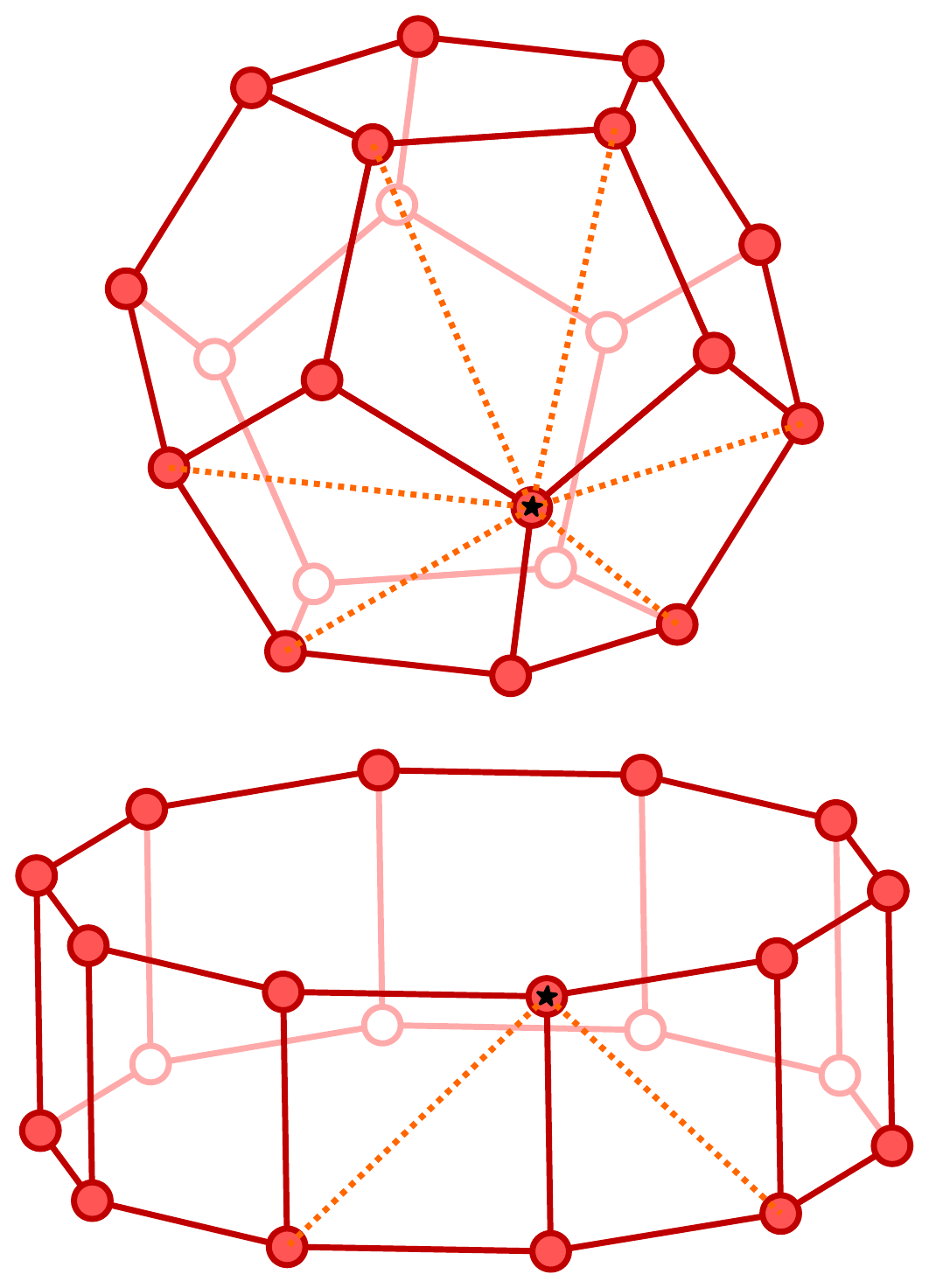}
    \caption{An example of a pair of point clouds which are clearly different, but have a graph that is indistinguishable by the Weisfeiler-Lehman test, if the graph is built based on a first-neighbor cutoff (full red lines). Increasing the cutoff to include the second neighbors (dashed orange lines) clearly allows discriminating between the two configurations.  }
    \label{fig:dodecaprism}
\end{figure}

Fig.~\ref{fig:dodecaprism} shows an example, adapted from Ref.~\citenum{sato2020arxiv}, of a pair of configurations whose distance-decorated graphs cannot be distinguished by a WL test \emph{provided that the graph only includes first-neighbor distances}. Several analogous examples have been shown for different types of CNNs \cite{zhan+21prl, vikas2020arxiv}, where the use of a finite cutoff in the construction of the molecular graph affects the resolving power of the network. 
As shown in the figure, however, increasing the cutoff distance to include further nodes in the definition of the graph is usually sufficient to distinguish the structures. In what follows, we will discuss a counterexample that cannot be resolved by simply using a bigger cutoff: a family of structures that fail the WL test even when considering their fully-connected molecular graph.

\begin{figure*}[bthp]
    \centering
    \includegraphics[width=0.9\linewidth]{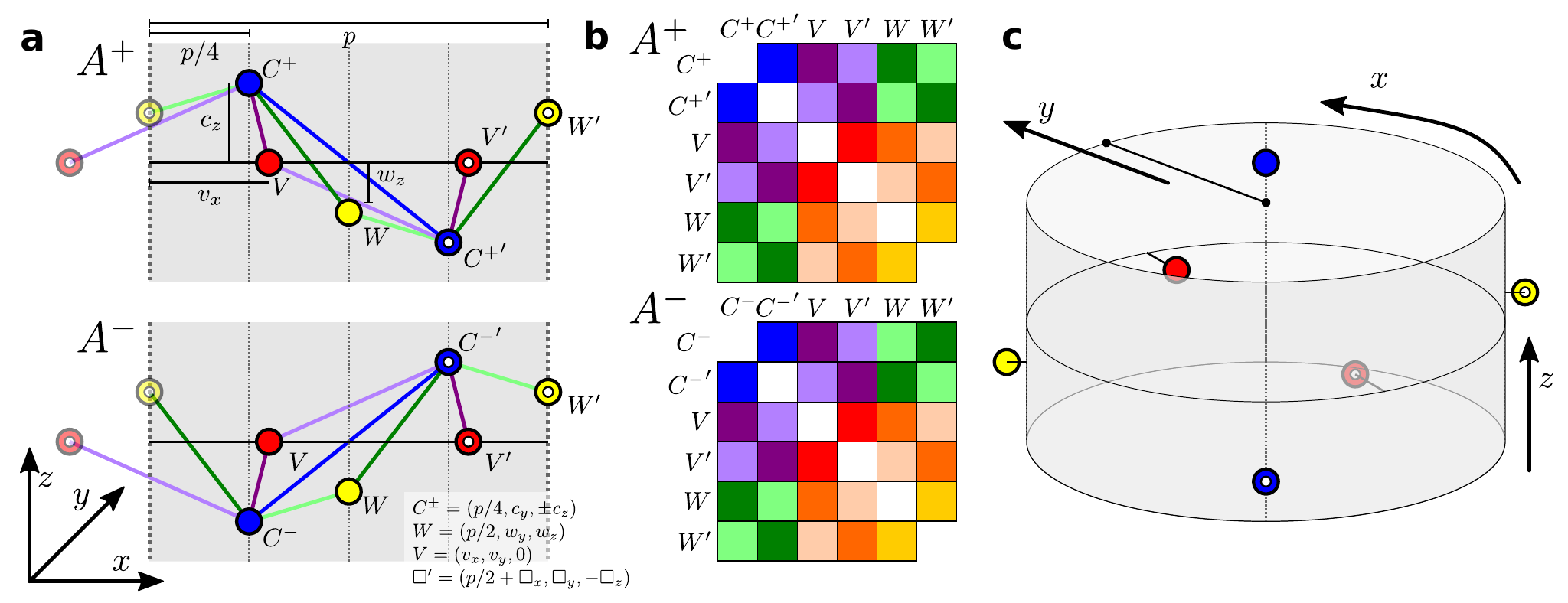}
    \caption{(a) Two structures, $A^+$, $A^-$ that generate pair-distance graphs that are indistinguishable based on a WL test. Both structures are periodic along $x$, and the coordinates of the six points are given in the figure and in Eq.~\ref{eq:coordinates}: the points have the same label in pairs; $V$ and $W$ points are identical in the two structures, while $C$ points are reflected relative to the $xy$ plane. Distances involving the $C^{\pm}$ points are highlighted, using a minimum image convention along the periodic direction. (b) Euclidean adjecency matrices between the points in the $A^+$ and $A^-$ structures. The two matrices differ only by the order of the $C-W$ distances. $V$ and $V'$ have the same set of edge distances, and so the $W/W'$ and $C^\pm/{C^\pm}'$ pairs: thus, from the point of view of the WL test, there are only three types of nodes and the graphs are indistinguishable. (c) A pair of finite-dimensional degenerate structures can be obtained by wrapping the period around the $z$ axis and embedding the periodic structure in Euclidean 3D space.  }
    \label{fig:cnn-scheme}
\end{figure*}

\section{A counterexample}

Consider the construction in Figure~\ref{fig:cnn-scheme}a. Six points, with labels that are identical in pairs, are arranged following the pattern in the figure, forming two structures, $A^+$ and $A^-$ with
\begin{equation}
\begin{array}{l}
C^\pm = (p/4, c_y, \pm c_z)\\
W = (p/2, w_y, w_z) \\
V = (v_x, v_y, 0) \\
\square ' = (p/2+\smallsquare_x,\smallsquare_y,-\smallsquare_z)\label{eq:coordinates}
\end{array}
\end{equation}
where the last definition indicates the relation between plain and primed points. 
The structures are periodic along the $x$ axis, with a period $p$, and have open boundary conditions along $y$ and $z$. After one iteration of the modified WL procedure discussed in Section~\ref{sec:w-l-test}, one sees that the unordered set of distances for $V/V'$, $W/W'$, $C/C'$ pairs are identical, so that the plain and primed points will receive the same hash value, and the two graphs cannot be discriminated by the test despite being different (Figure~\ref{fig:cnn-scheme}b).
The essential ingredient to induce a degeneracy is the swapping of the distances between $C$ and $W$ points, i.e. that $\dst(C^+,W)=\dst(C^-,W')$ and $\dst(C^+,W')=\dst(C^-,W)$, { which changes the geometry of the structures but does not affect the outcome of the WL test, that relies only on the unordered set of distances. This swap, which can be clearly seen by plotting the distance matrix, is a consequence of the regular spacing of the points along the $x$ axis and the periodic boundary conditions. Distances between the $C$ points and the $W$ points to their left in the $A^+$ structure correspond to distances between the $C$ points and the $W$ points to their right in the $A^-$ configuration. A more thorough discussion of the role of periodicity is also given in Appendix~\ref{app:periodic-3d}, which deals with the 3D-periodic case.}

\begin{figure}[bp]
\centering
   \includegraphics[width=0.95\linewidth]{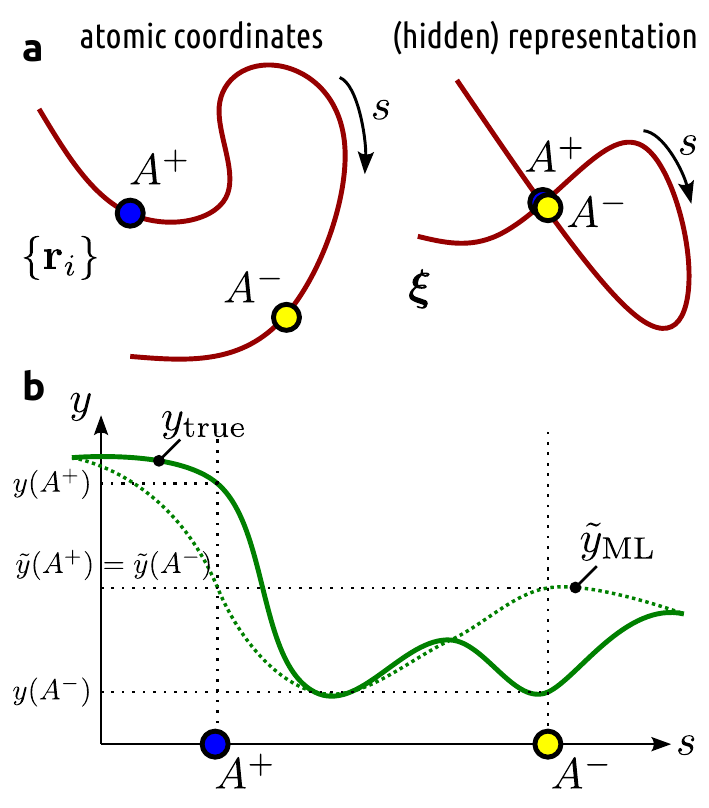} 
   \caption{\rev{(a) A cartoon depicting a representative path in coordinate space that joins a degenerate pair $A^\pm$ (left) and the deformation it induces in the space of features/hidden representation of an incomplete model. (b) The geometric degeneracy implies that the model makes the same prediction for $\tilde{y}(A^+)$ and $\tilde{y}(A^+)$. Smoothness implies that nearby structures that are not strictly degenerate are also affected. \label{fig:explained} }}
\end{figure}

It is important to stress that even though we only show minimum-image distances in the figure, the graphs generated by the periodic structure are degenerate even when considering a fully-connected graph, including all distances between periodic replicas. 
Thus, the effectiveness of the counterexample is independent on the details of the definition of the graph neighborhoods, and it equally affects cases in which one only considers the neighbors within a fixed cutoff distance, or just the first $k$ neighbors. 
{The WL procedure is a very powerful test to differentiate graphs based on first-order connectivity information. }
Failure in distinguishing graphs based on the WL test indicates that the pairs are equivalent for a broad class of pair-distance GNNs, \cite{gilm+17icml} that includes SchNet \cite{schu+18jcp}, convolutional-networks  molecular fingerprints \cite{duve+15nips}, molecular graphs convolutions \cite{kear+16jcamd}, graph networks \cite{chen+19cm}, and many others, {which from a graph theoretical perspective form a hierarchy of tests, all less powerful than the WL analysis.\cite{xu2018how}}
As a special case, the $A^\pm$ structures in Fig.~\ref{fig:cnn-scheme}a are also degenerate with respect to the \emph{full set} of neighbor-distances multisets, which means they are an example for which $\nu=1$ atom-centered features cannot discriminate \emph{globally} between the two structures. In this sense, this is a a much more difficult case than the degenerate tetrahedra in Fig.~\ref{fig:classics}a, which have the same set of distances but can be easily distinguished when considering the triplets of distances  associated with the vertices, and of that in Fig.~\ref{fig:classics}b, which is globally undistinguishable by $\nu=1$ representations, but can be resolved by a dGNN.

This construction produces a pair of continuous manifolds of dimension 7 (and co-dimension 8, discarding translations), and can be further generalized in several different ways. Arbitrarily many pairs of $V$ and $W$ atoms, possibly with different labels, can be added to the two structures without breaking the degeneracy. { Each pair increases the dimension of the manifold by two and the codimension by four. Structures of any size and complexity can be generated with this construction, even though the increase in co-dimension suggests that they become less ``dense''.} 
\rev{
One should keep in mind that the presence of degenerate configurations affects the accuracy and numerical stability of ML models even for other structures.\cite{pars-goed22jcp,pozd+22jcp} 
From the point of view of the input features (or more broadly, the hidden representation of a deep-learning framework), an overlap of two structures that should be distinct determines a distortion that brings close together structures that should be far apart (Fig.~\ref{fig:explained} a). 
From the point of view of predicting properties, e.g. the potential energy of a molecule, the smoothness of the approximation is a key requirement to ensure stability and transferability of the model, which is fulfilled by all practical implementations. 
Thus, small deformations of a molecule should result in small changes of the target, and so a degenerate model will predict similar values for all structures that are close to a pair of degenerate configurations, even though they are not exactly degenerate (Fig.~\ref{fig:explained}b). 
}

{As we discuss in Appendix~\ref{app:periodic-3d}, periodic boundary conditions can be added to the $y$ and $z$ axis without lifting the degeneracy of the pairs. }
Intriguingly, it is also possible to ``fold'' an arbitrary number $P$ of repeat units of the structure around the $z$ axis, to obtain a pair of finite structures embedded in 3D Euclidean space (Figure~\ref{fig:cnn-scheme}c). 
If the coordinates in the $x$-periodic structure are $\{(x_i,y_i, z_i)\}$, the position of the finite-size structure are
\begin{equation}
\begin{array}{ll}
r_i = & P p/2\pi + y_i\\
\theta_i = & 2\pi x_i /p P \\
z_i = & z_i 
\end{array}\label{eq:finite-size}
\end{equation}
in a cylindrical coordinate system. This coordinate transformation works because it maps equal distances in the periodic structures to equal Euclidean distances in the finite configurations, and so it does not affect the nature of the graphs, and their signature when subject to the WL test. 

{
\section{Beyond distance-based GNNs}

All of these structures are easily distinguished by incorporating information on the angles, so that any scheme that contains at least 3-body descriptors, such as SOAP features \cite{bart+13prb}, atom-centered symmetry functions \cite{behl-parr07prl}, are not affected by this counterexample. {As a consequence, higher-order graph neural networks \cite{morr+18arxiv}, as well as many other deep learning approaches that incorporate angular information (such as DimeNET\cite{Gasteiger2020Directional}
GemNET\cite{klic+21arxiv}, REANN\cite{10.1063/5.0080766}, ALIGNN\cite{chou-deco21npjcm}, etc.) are also immune to these counterexamples, even though one cannot exclude that 3D counterexamples may be found also for some of these architectures, given that it is known that general graphs that defy higher-order versions of the WL test exist.} 
{
Frameworks that can generate systematically arbitrary high-order correlation features\cite{will+19jcp} (such as MTP\cite{shap16mms}, ACE~\cite{drau19prb}, NICE~\cite{niga+20jcp}, etc.)  can be shown to be provide a complete description of interatomic correlations \cite{sanc+84pa,musi+21cr} and are therefore, in the appropriate limit, symmetry-adapted universal approximators \cite{dym-maro20arxiv}.
Similarly, equivariant neural networks (such as Tensor Field networks\cite{thom+18arxiv}, Cormorant\cite{ande+19nips}, etc.) incorporate high-order correlation information by combining information on the interatomic distance \emph{vectors}, constraining the functional form of the messages in a way that preserves equivariance, and achieve completeness in a very similar way as for the high-order correlation features\cite{niga+22jcp2}. }

The emergence of models which, in theory, have universal interpolating properties does not make the search for this kind of pathological configurations less important. 
For example, they can help verify whether the practical implementation of a ML scheme is consistent with its theoretical properties. 
Take for instance the universal scalar framework of Ref.~\citenum{vill+21nips}.
A universal approximator for vectorial functions of the coordinates of $n_\text{atoms}$ particles, that are equivariant in O(3) and invariant with respect to atom index permutations is proposed in the form (Eq. 11 in the reference)
\begin{equation}
\sum_{t=1}^{n_\text{atoms}}
f(\mathbf{x}_t, \mathbf{x}_1, \ldots \mathbf{x}_{t-1},\mathbf{x}_{t+1},\ldots, \mathbf{x}_{n_\text{atoms}}) 
\mathbf{x}_t,
\end{equation}
where $f$ is an arbitrary scalar function that is O(3) invariant, and invariant to the permutation of its arguments past the first. 
The crux of problem, however, is how to implement in practice such a universal symmetric scalar function. 
The example that is provided in section 7 of Ref.~\citenum{vill+21nips}  uses a functional form that is suitable to learn a restricted class of targets, but is \emph{not} a universal approximator. The proof for the full functional form is rather cumbersome, and we report it in Appendix~\ref{app:scalars-universal}, but the key problem is that the scalar functions are written in terms of the multi-set of the elements of the Gram matrix, e.g.
$f(\{\mathbf{x}_i\cdot \mathbf{x}_j \}_{i,j=1}^{n_\text{atoms}})$, and therefore cannot distinguish between the degenerate structures we introduce here, and neither between the planar configurations in Fig.~\ref{fig:classics}c, that only differ by the order of the entries of their Gram matrix.

Furthermore, provably universal equivariant frameworks are such in the limit in which they generate high-order correlations -- either explicitly or as a consequence of the stacking of equivariant layers\cite{niga+22jcp2}. 
It is an interesting, and open, question whether a given order suffices to guarantee complete resolving power. There is no reason to believe that the present family of degenerate structures is the \emph{only} one that could be realized, based on a similar idea of generating configurations in which some distance pairs are swapped.
It remains to be seen if a combination of the ideas we introduce here with those that underlie the counterexamples\cite{pozd+20prl} for atom-centered descriptors of order $\nu=2,3$ could allow constructing configurations that are indistinguishable to convolutional schemes that also use angular (and possibly dihedral) information. }

\begin{figure}[tb]
    \centering
    \includegraphics[width=1.0\linewidth]{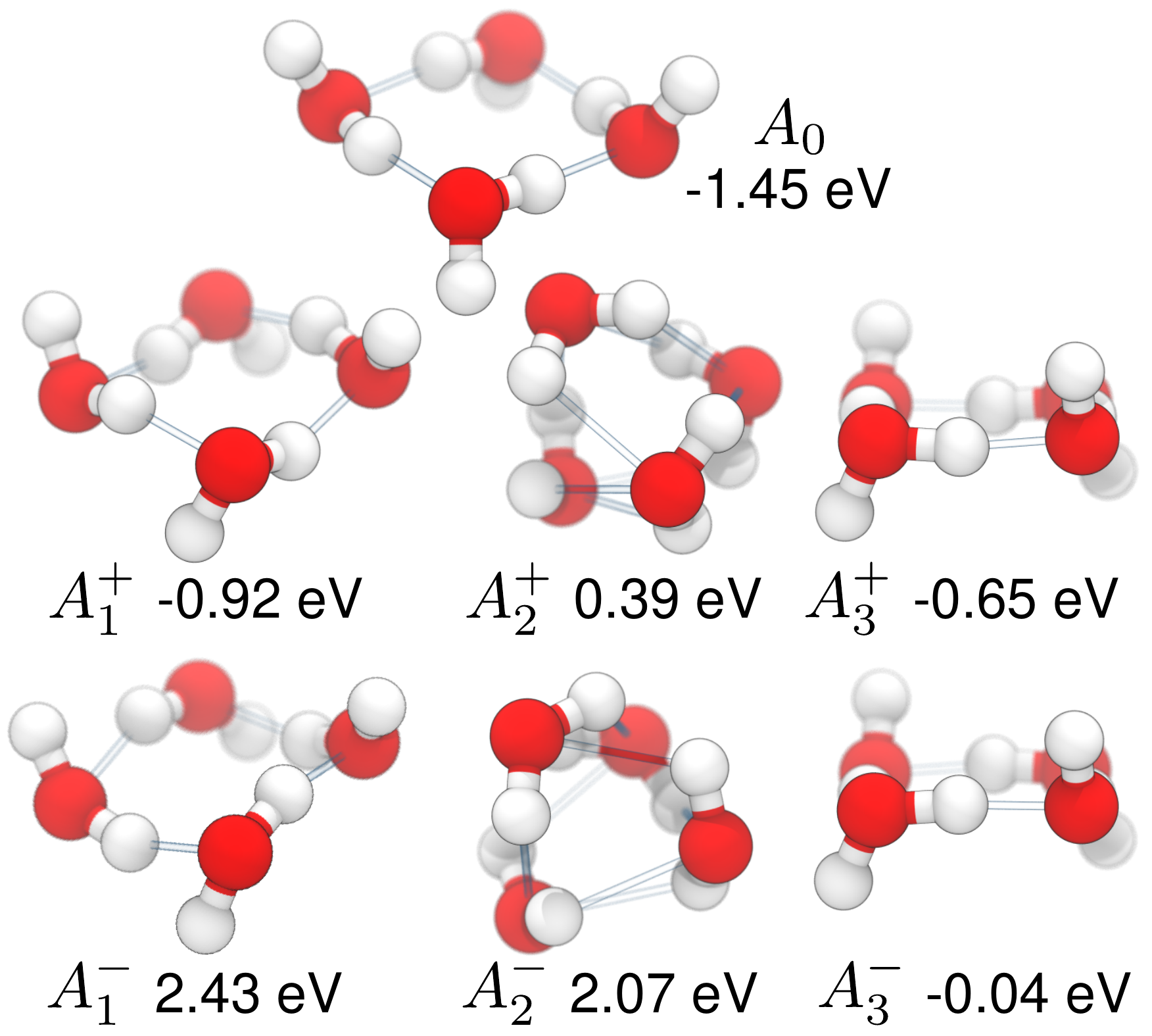}
    \caption{A few configurations of a water tetramer that will be used as benchmarks. $A_0$ is the ground state structure for the tetramer \cite{wale-wals97jcp}. $A_1^\pm$, $A_2^\pm$ and $A_3^\pm$ are three pairs of structures that correspond to two periods of the 3D motif of Fig.~\ref{fig:cnn-scheme}c.
    The geometries of the $A^{+}_{1,2}$ structures have been obtained by a constrained minimization of the energy, while $A^{\pm}_3$ have been obtained by minimizing simultaneously the energies of both degenerate structures. }
    \label{fig:tetramers}
\end{figure}

\section{Significance for chemical ML}

\begin{figure}[tb]
    \centering
    \includegraphics[width=1.0\linewidth]{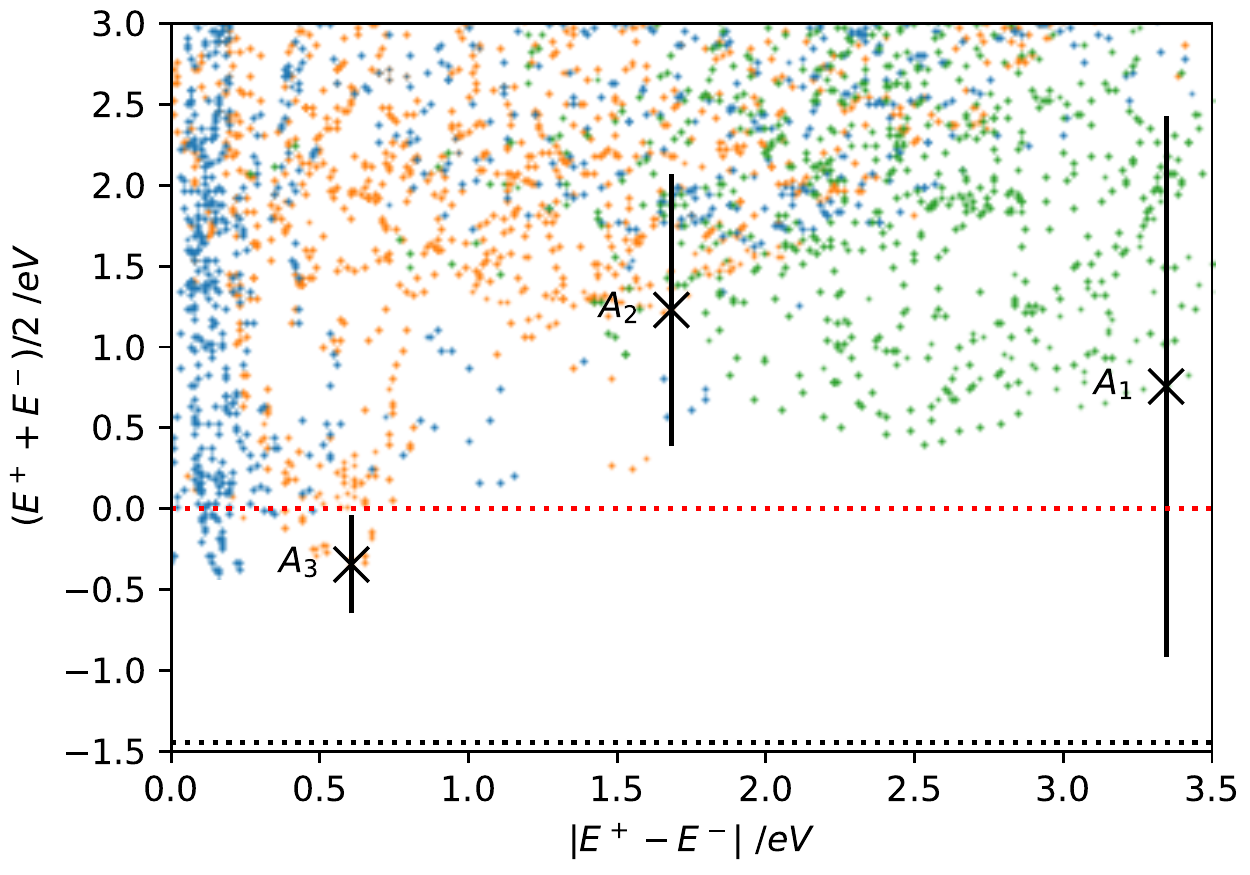}
    \caption{
Scatter plot showing the energies (relative to the fully dissociated \ce{4H2O} geometry) of several hundreds of cluster geometries, obtained by finite changes of the parameters of the three structures shown in Fig.~\ref{fig:tetramers}. 
For each pair of degenerate conformers, a point is shown indicating the mean and the spread of the two energies. 
Points corresponding to $A_{1,2,3}$ are shown as black crosses, and a bar indicates the energies of the two structures in the pair. 
The dissociation limit and the energy of the ground-state structure are shown as red and black dashed lines.}
    \label{fig:clouds}
\end{figure}

The construction in Fig.~\ref{fig:cnn-scheme} might seem somewhat contrived, but it is not difficult to use it to generate point clouds that correspond to realistic targets for machine learning. Consider for example the case of predicting the stability of molecular structures. Fig.~\ref{fig:tetramers} demonstrates three pairs of degenerate configurations corresponding to a water tetramer, {that we obtained based on the expression for a finite structure~\eqref{eq:finite-size} with two repeat units.} 
Structure $A_1^+$, in particular, is a mildly distorted version of the ground state configuration of \ce{(H2O)4},\cite{mahe+01jpca} that we label as $A_0$.
The energies of these structures (which we computed at the B3LYP \cite{beck93jcp} level, using PySCF \cite{sun+20jcp}) are not absurd: the cohesive energies relative to four isolated \ce{H2O} molecules are $E_1^+=$-0.92~eV, $E_1^-=$2.43~eV, $E_2^+=$0.39~eV, $E_2^-=$2.07~eV. 
These structures are part of a continuous manifold of degenerate pairs: as shown in Fig.~\ref{fig:clouds}, simply modifying the parameters in the  construction~\eqref{eq:coordinates} around these specific values generates hundreds of configurations with energies comparable with the dissociation energy of the tetramer. 
In particular, one can also obtain pairs for which both structures are below the cluster dissociation energy: we consider as an example the $A_3^\pm$ structures, which have $E_3^+=$-0.65~eV, $E_3^-=$-0.04~eV.

Thus, in the best-case scenario a distance-based GNN can only predict the energies of $A_{1,2,3}^{\pm}$ with an RMSE of $\approx 1.1$~eV. For reference, the cohesive energy of the ground-state structure is $E_0=-1.45$~eV, and the typical thermal energy at room temperature is $0.025$~eV.
Water oligomers have been extensively studied as a way to understand the properties of the hydrogen bond network in water \cite{liu+96science,zwie04science}, that in turn influences the behavior of all chemical and biochemical phenomena that take place in solution.
From a modelling point of view, an accurate estimation of the energetics of 2 and 3-molecule clusters is important because it underlies the construction of very accurate interatomic models of the energetics of bulk water, such as the MB-pol potential \cite{medd+15jcp,nguy+18jcp} that adds to a polarizable baseline that captures long-range interactions a series of short-range corrections based on fitting to high-end quantum calculations of dimers and trimers. \rev{These terms are then computed by summing over all pairs or triplets of water molecules in the system. Even for a bulk configuration, one has to evaluate cluster energies.  }

In this context, the example we present is very relevant, in view of the growing interest in incorporating explicitly 4-molecule terms \cite{hein-xant21jctc}.
With this application in mind, we chose the data set used in training the MB-pol potential \cite{medd+15jcp,nguy+18jcp} to train models of the cluster cohesive energy,
selecting 10'000 water dimers and 5'000 water trimers by farthest-point sampling (FPS)\cite{imba+18jcp}, combined with 2'000 water tetramers configurations (FPS selected from a high-temperature Hamiltonian replica exchange simulation with a confining potential, performed using i-PI \cite{kapi+19cpc} and the q-TIP4P/f forcefield \cite{habe+09jcp}). 
Energies of all structures were then computed with the same B3LYP setup as for the degenerate structures. 

\begin{figure}[tb]
    \centering
    \includegraphics[width=1.0\linewidth]{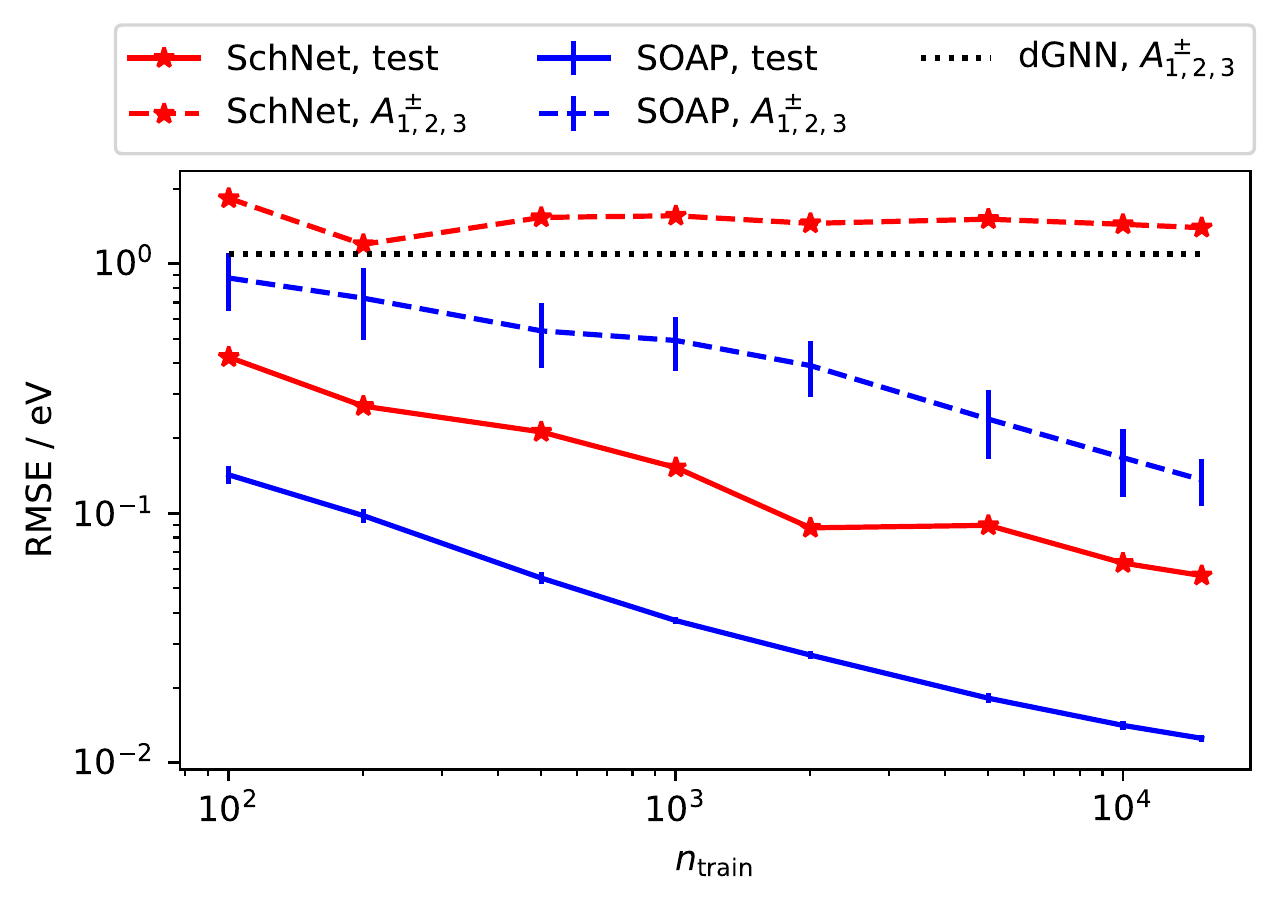}
    \caption{Learning curves for SchNet (red lines) and SOAP GAP (blue lines) models trained on a dataset of 2-, 3-, and 4-water clusters. Full lines show validation error on a hold-out set of 2'000 structures, dashed lines the RMSE for the four  $A^{+}_{1,2}$ geometries. 
    The dotted black line corresponds to the ultimate limit for a dGNN that cannot discriminate between the degenerate pairs. }
    \label{fig:tetramer-models}
\end{figure}

We show a comparison between a simple kernel model based on SOAP features \cite{bart+13prb} and the SchNet framework \cite{schu+17nips}, a GNN that has been very successfully applied to several chemical ML problems \cite{schu+19nc,bror19jcp,dand+20jpca,west+20jpcl,west-maur21cs}.
In SchNet \cite{schu+17nips}, each atomic environment $A_i$ is associated with a feature vector $\bfeat(A_i)$, that is initialized to values that depend only on the chemical nature of the $i$-th atom. These are updated with on-site operations, and with interaction blocks that combine information on each of the neighbors according to
\begin{equation}
\feat^{k+1}_q(A_i) = \sum_{j\in A_i} \feat^{k}_q(A_j) W_q(r_{ij}),
\end{equation}
where $W_q(r)$ indicates a set of distance-dependent continuous filters, and $\feat^{k}_q(A_j)$ indicates the features of the neighbors from the previous iteration of graph convolution. 
This functional form incorporates only the type of information associated with the distance-decorated WL test: the attributes of the neighbors, and scalar attributes of the interatomic separation, that are combined in a permutation-invariant manner.
SOAP features, on the other hand, contain information on the angular relations between neighbors. Even though we evaluate them by first computing an expansion of the neighbor density on spherical harmonics, and then contracting the expansion coefficients to extract the rotationally-invariant components \cite{bart+13prb}, SOAP features can also be expressed in a form that highlights their dependence on neighbor-neighbor angles
\begin{equation}
\feat_q(A_i) = \sum_{j,j' \in A_i} W_q(r_{ij}, r_{ij'}, \br_{ij}\cdot\br_{ij'}).
\end{equation}
We combine these features to build polynomial kernels $\krn(A_i,A_{i'})=(\bfeat(A_i)\cdot \bfeat(A_{i'}))^4$, that are then used for kernel ridge regression.
We emphasize that we chose SchNet as a widespread GNN model, but \emph{any} distance-based GNN that has a discriminating power equivalent to (or lower than) a first-order WL test would exhibit the same problem.
Similarly, we use a SOAP-based Gaussian Approximation Potential as a simple and well-understood scheme that incorporates explicitly angular information, but essentially any framework that does so would be capable of resolving the $A^{\pm}$ degeneracies.

We use standard hyperparameters for both SchNet and the SOAP model (example scripts, full model parameters and the training and test sets are provided as supporting materials). As shown in Fig.~\ref{fig:tetramer-models}, the difference in behavior between the two schemes is qualitative. 
Both models can reduce monotonically the validation error on the $n$-mers dataset, but only the model that incorporates angular information can tell the $A^\pm$ pairs apart, and can bring the errors for $A^\pm_{1,2,3}$ below the theoretical limit that corresponds to predicting for both structures the mean of $E^+$ and $E^-$. 
The figure also shows that the two models exhibit very different performances, and that the SOAP-based kernel model yields a much larger error for $A^\pm_{1,2,3}$ than for the test set, which does contain structures with a similar energy range. 
It is difficult to conclusively determine the reason for these observations, that depend at least in part on the detailed setup of the two models, which we have deliberately not optimized, because the qualitative observation of the failure of dGNNs is independent on parameters or implementation details. 
One may hypothesize that structures that are only distinguishable by angular information may be more challenging for a SOAP model -- because of the higher complexity of a 3-body potential compared to pair terms. As for the dGNNs, even in the absence of structures that are \emph{exactly} degenerate, one can expect that the lack of resolving power is reflected in a slower convergence. 
Structures that lie in the vicinity of a singularity can be distinguished by the dGNN, but only by sacrificing the smoothness, and hence the data-efficiency, of the approximation \rev{(see also Fig.~\ref{fig:explained}). This argument provides a plausible, although not conclusive, explanation for the improvement in performance that is observed when comparing dGNNs with frameworks that incorporate angular or directional information\cite{pmlr-v139-schutt21a}, and is consistent with similar considerations that can be made to explain the saturation of learning curves for SOAP-based models applied to a dataset containing structures close to two-neighbors degeneracies (Fig.~\ref{fig:classics}c). \cite{pozd+20prl} }

\begin{figure}[bt]
    \centering
    \includegraphics[width=1.0\linewidth]{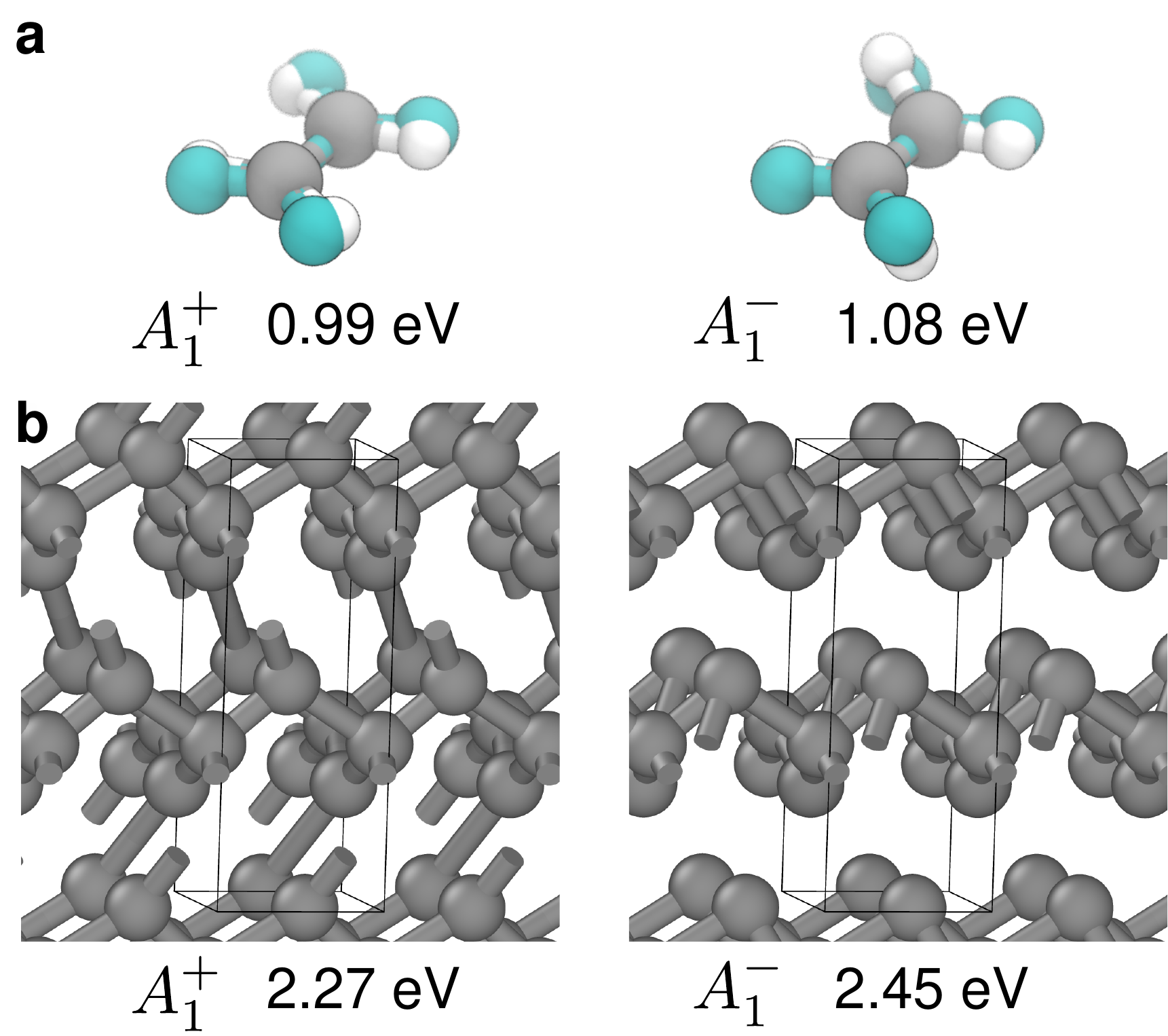}
    \caption{{(a) A pair of configurations of ethene, \ce{C2H4}, that cannot be distinguished by a dGNN. The structures are 1eV less stable than the minimum energy configuration (represented in cyan to visualize more clearly the distortion). 
    (b) A pair of 3D-periodic structures which cannot be distinguished by a dGNN. When interpreted as being composed by carbon atoms, their energies as computed by DFT in the local density approximation lie about 2eV/atom above the energy of cubic diamond. }}
    \label{fig:carbon}
\end{figure}

{We conclude by reiterating that this is just one of many possible examples that can be realized using our construction. 
Figure~\ref{fig:carbon} shows a pair of ethene configurations that cannot be distinguished by a WL test. What is more, it also shows a pair of bulk carbon structures that are indistinguishable. The existence of 3D periodic structures that are indistinguishable even when using an infinite distance cutoff  provides a clear demonstration that the degeneracies are not simply explained by the presence of a small number of interatomic distances. 
Rather, they are associated with the existence of periodicity and near-symmetries, so that the distances that decorate the fully-connected graph contain redundant information that is not sufficient to distinguish $A^+$ from $A^-$.
We emphasize that we provide these structures as examples of configurations which defy dGNNs, and as a test to verify whether a given model can resolve the counterexamples we discuss here. There is at least one continuous manifold of degenerate geometries, and it is likely that a more thorough search, including the use of different chemical species, would lead to conformers that are even closer in energy to the thermodynamically-stable conformations and phases.
 }

\section{Conclusions}

The idea of combining a characterization of point clouds in terms of graphs, where points are vertices and the connectivity is determined by point-to-point distances, with graph-convolution architectures has been extraordinarily successful in a broad range of applications of geometric machine learning, and in particular in the construction of deep models of chemical structures, that lend themselves naturally to such description. 
Despite their simplicity, and despite the fact that general graphs that are indistinguishable based on first-order GNN are known to exist, these frameworks have remarkable descriptive power for actual molecular structures, that correspond to a special class of fully-connected graphs with edges decorated by the distances between the atoms. For example, they are capable of discriminating between configurations that are degenerate to widely used atom-centered representations. The previously known cases in which they fail to distinguish two structures depended on the choice of a small, finite cutoff in the definition of the molecular graph. 
The construction we present here, that generates pairs of geometries that are indistinguishable when seen through the lens of a dGNN, regardless of the chosen cutoff, provides a counterexample that sets a limit to the accuracy that can be attained when regressing properties associated with the 3D point cloud. 
For the specific case of chemical machine learning, we show a concrete case, relevant to the  chemistry of water clusters, demonstrating that the counterexample can have direct repercussions on practical applications, such as the training of many-body potentials for water and aqueous systems.  

Atom-centered descriptors that rely on correlations with at least two neighbors, {such as SOAP,\cite{bart+13prb} Behler-Parrinello symmetry functions,\cite{behl11jcp}} etc.  -- as well as high-order graph convolution schemes and equivariant neural networks -- are immune to this family of counterexamples, \rev{although low-order correlations are affected by other types of degeneracies. Atom-centerd descriptors of higher order, or equivariant network architectures,  can in principle be made complete, often resulting in improved performance on chemical learning tasks. }
Determining rigorously the \emph{minimal} amount of information that must be incorporated in geometric machine learning to guarantee that an architecture can resolve arbitrary point clouds is a fascinating topic that touches upon several open problems in signal processing \cite{kaka12jmiv} and invariant theory.\cite{bout-kemp04aam}
We hope that our data set will be used to test existing frameworks to determine empirically their descriptive power, and that our construction could be taken as a basis to break some of the angle-dependent schemes and to determine the simplest architecture that is not affected by this, or other, degeneracies.

\begin{acknowledgments}

MC and SNP acknowledge support from the Swiss National Science Foundation (Project No. 200021-182057), from the Platform for Advanced Scientific Computing
and from the NCCR MARVEL, funded by the Swiss National Science Foundation (grant number 182892). The authors gratefully acknowledge discussion Bin Jiang and Yaolong Zhang, Rose Cersonsky for suggestions on data visualization, and Vitaliy Kurlin for suggesting to extend the periodic counterexample to the 3D case.

\end{acknowledgments}

\appendix

\begin{figure}[b]
\centering
\includegraphics[width=1.0\linewidth]{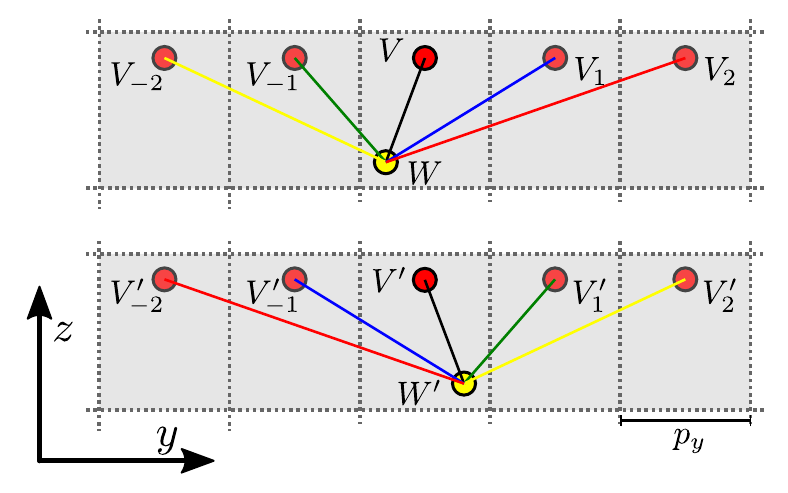}
\caption{{A 2D example demonstrating that displacement vectors with opposite-sign components lead to the same sets of distances between periodic replicas. 
The displacement vectors $\protect\overrightarrow{VW}$ and $\protect\overrightarrow{V'W'}$ relate to each other as $(\Delta y, \Delta z)=(-\Delta y, \Delta z)$, and as a consequence the displacement vectors with periodic replicas along $y$ can be put in one-to-one correspondence (specifically, $\dst(W,V_i)=\dst(W',V'_{-i})$),
and the unordered sets of all the distances to the periodic replicas of $V$ and $V'$ are the same. The same would be true along the $z$ direction, and (in the 3D case) along the $x$ direction. } } \label{fig:permutation}
\end{figure}

{
\section{Extension of the counterexample to 2D and 3D periodic structures}\label{app:periodic-3d}

One of the reasons why the pairs of configurations $(A^+,A^-)$ we introduce in this work are indistinguishable to a WL test is the fact that each atomic environment in structure $A^+$ has a corresponding environment in $A^-$ with the same set of neighbor distances. 
Introducing additional neighbors (as one does when considering periodicity along $y$ and/or $z$) increases the size of the graph and the number of interatomic distances. This additional information could break the degeneracy of one or more environments. 

In general, the equality of a pair of distances does not ensure that the distances corresponding to periodic replicas of the atoms involved will be the same: $\|(\Delta x_1, \Delta y_1, \Delta z_1)\|^2 =\|(\Delta x_2, \Delta y_2, \Delta z_2)\|^2  $ does not guarantee that 
\begin{multline}
\|(\Delta x_1 +n_x p_x, \Delta y_1 +n_y p_y, \Delta z_1 +n_z p_z)\|^2 \\ =\|(\Delta x_2+n_x p_x, \Delta y_2 + n_y p_y, \Delta z_2 +n_z p_z)\|^2 
\label{eq:replica-distances}
\end{multline}
for all periodicities $(p_x,p_y,p_z)$ and cell indices $(n_x,n_y,n_z)$. 
A special case for which Eq.~\eqref{eq:replica-distances} clearly holds is that in which not just the distances, but the full \emph{distance vectors} are equal, $(\Delta x_1, \Delta y_1, \Delta z_1) = (\Delta x_2, \Delta y_2, \Delta z_2)$. This is the case for several distance pairs in the structures in Fig.~\ref{fig:clouds} (e.g. the displacement vector  $\overrightarrow{VC^+}$ equals $\overrightarrow{V'{C^-}'}$) but would not suffice, alone, to ensure that the fully-periodic structures are degenerate. 
We must consider an additional, more subtle case, in which the distance vectors have one or more components which are equal in magnitude, but have opposite signs, e.g. $\Delta z_1  = - \Delta z_2$. In this case, while it is not true that the distances corresponding to the same replica  $(n_x,n_y,n_z)$ will be equal, it will be possible to establish a 1-to-1 mapping between the replicas -- in this case,  $(n_x,n_y,n_z)\rightarrow  (n_x,n_y,-n_z)$ (see Fig.~\ref{fig:permutation}), and so the unordered sets of distances will be equal.

One can then see that in the original 1D construction illustrated in Fig. \ref{fig:cnn-scheme}a, every time distances between pairs of points are the same (i.e., are marked by the same color in the figure), then the corresponding displacement vectors consist of the same Cartesian components with, possibly, opposite signs. According to the discussion above, this implies that if some distances within the original 1D counterexample are the same, then the unordered sets of distances associated with the replicas along the $y$ and/or $z$ axis are also the same. Thus, the WL test (which operates exclusively based on the unordered sets of distances) cannot discriminate between the 2D and 3D analogues of the 1D construction. 
}

\section{Counterexample for a ``universal approximator'' implementation}
\label{app:scalars-universal}

Ref.~\citenum{vill+21nips} proposes a general recipe to construct universal approximators on point clouds. As a specific example, the authors propose learning a function defined for a point cloud described by $N$ 3D coordinates $\br_i$ and a scalar node property $m_i$ (interpreted as a point mass). 
The goal is to learn a rotationally covariant and permutationally invariant tensorial function $h$: 
\begin{equation}
h\: : \: (\mathbb{R} \times \mathbb{R}^3)^N \rightarrow \mathbb{R}^{3\times3}.
\label{eq:h-form}
\end{equation}
The authors propose a functional form of the type 
\begin{multline}
\label{eq:suggested}
h((m_i, \br_i)_{i=1}^N) = \\\sum_{i=1}^N f_0(\br_i^T \br_i, m_i, \{\br_k^T \br_l, m_k, m_l\}_{k,l\neq i}) \br_i \br_i^T \\
+ \sum_{i>j=1}^N f_1(\br_i^T\br_j, m_i, m_j, \{\br_k^T\br_l, m_k, m_l\}_{k,l\neq i,j}) \br_i \br_j^T \\
+ f_2(\{\br_i^T\br_j, m_i, m_j\}_{i,j=1}^N)\mathbf{1}
\end{multline}
where $\mathbf{1}$ indicates the identity matrix and $\{...\}$ indicates unordered multisets. We are interested in determining whether this form provides a universal approximator for symmetric functions of the type~\ref{eq:h-form}, and present a counterexample showing that this is not the case. 

We use a simpler geometry than the general construction we present in the main text, but the counterexample is based on similar principles. 
This pair of structures is shown in Fig.~\ref{fig:classics}b, and is also a counterexample for 2-body atom centered additive models. 
We chose this pair because the coordinates take integer values, and so it is simpler to show explicit values of the various quantities.  
The general case would also result in the impossibility of learning the targets. We also take all $m_i$ equal to each other, so that the $m_i$ can be dropped from the definition~\ref{eq:suggested}.
The coordinates of the first set of points are given by:
\begin{equation}
    (\br_i^+)_{i=1}^5 = \begin{bmatrix}
    1  & -1  & 2  & -2  & 0  \\
1  & -1  & 0  & 0  & 1  \\
0  & 0  & 2  & -2  & 1
    \end{bmatrix}^T
\end{equation}
while those of the second set of points are
\begin{equation}
    (\br_i^-)_{i=1}^5= \begin{bmatrix}
    1  & -1  & 2  & -2  & 0  \\
1  & -1  & 0  & 0  & 1  \\
0  & 0  & 2  & -2  & -1  
    \end{bmatrix}^T
\end{equation}
We use a simple diagonal function as target, 
\begin{equation}
 h((\br_i)_{i=1}^5) = \sum_k \lambda_k^3 \mathbf{1},
\end{equation}
which is constructed by summing over the eigenvalues $\lambda_k$ of the Gram matrix of each structure, and is clearly rotationally covariant and invariant with respect to permutations of the points in the point cloud. The value of the function $h$ for the two point clouds differ by
\begin{equation}
    h((\br_i^+)_{i=1}^5)  -  h((\br_i^-)_{i=1}^5)= \begin{bmatrix}
192 & 0 & \\
0 & 192 & 0 \\
0 & 0 & 192
\end{bmatrix}.
\end{equation}

We decompose the difference between the predictions given by the form \ref{eq:suggested} as:
\begin{equation}
 h((\br_i^+)_{i=1}^5) - h((\br_i^-)_{i=1}^5)  = \Delta_0 + \Delta_1 + \Delta_2,
\end{equation}
where each $\Delta_k$ corresponds to the difference between the values of the term that contains $f_k$. 
The entries of the Gram matrix of the two structures are the same (even though ordered differently, so that their eigenvalues differ), and so obviously $\Delta_2$ has to be zero, given that $f_2$ is evaluated on identical sets. 
The expression for $\Delta_0$ contains 10 terms, but most cancel out, leaving just 
\begin{equation}
\begin{split}
\Delta_0 = & f_0(\zeta)  
\begin{bmatrix}
0 & 0 & 0 \\
0 & 0 & 2 \\
0 & 2 & 0
\end{bmatrix} =  \begin{bmatrix}
0 & 0 & 0 \\
0 & 0 & 2 f_0(\zeta)\\
0 & 2f_0(\zeta) & 0
\end{bmatrix},\\
\zeta =& (2, \{-8, -8, -2, -2, -2, -2, -2, -2,  \\
& \quad 2, 2, 2, 2, 2, 2, 8, 8\}).
\end{split}
\end{equation}
Similarly, the expression for $\Delta_1$ contains 20 terms, but most cancel out, leaving an expression of the form
\begin{equation}
\begin{split}    
\Delta_1 = &  \begin{bmatrix}
    0 & -4 f_1(\kappa_1) + 4 f_1(\kappa_2) & 2 f_1(\kappa_3) - 2 f_1(\kappa_4) \\
    0 & 0 & 2 f_1(\kappa_3) - 2 f_1(\kappa_4) \\
    0 & -4 f_1(\kappa_1) + 4 f_1(\kappa_2) & 0
    \end{bmatrix} \\
\kappa_1 = & (-2, \{-8, -8, -2, -2, -2, -2, -2, -2, -2, \\&\quad -1, -1, 1, 1, 2, 2, 2, 2, 2, 2, 2, 2, 2, 8, 8\}) \\
\kappa_2 = & (2, \{-8, -8, -2, -2, -2, -2, -2, -2, -2, -2, \\&\quad -1, -1, 1, 1, 2, 2, 2, 2, 2, 2, 2, 2, 8, 8\}) \\
\kappa_3 = & (1, \{-8, -8, -2, -2, -2, -2, -2, -2, -2, -2, \\&\quad -1, -1, 1, 2, 2, 2, 2, 2, 2, 2, 2, 2, 8, 8\}) \\
\kappa_4 = & (-1, \{-8, -8, -2, -2, -2, -2, -2, -2, -2, -2,\\&\quad  -1, 1, 1, 2, 2, 2, 2, 2, 2, 2, 2, 2, 8, 8\}) 
\end{split}    
\end{equation}

Combining the different parts, we obtain that 
\begin{multline}
h((\br_i^+)_{i=1}^5) - h((\br_i^-)_{i=1}^5) = \\
  \begin{bmatrix}
    0 & 4 (f_1(\kappa_2) -f_1(\kappa_1)) & 2 (f_1(\kappa_3) - f_1(\kappa_4)) \\
    0 & 0 & 2 (f_1(\kappa_3) - f_1(\kappa_4) + f_0(\zeta)) \\
    0 & 4 (f_1(\kappa_2) -f_1(\kappa_1)) + 2 f_0(\zeta) & 0
    \end{bmatrix},
\end{multline}    
which is incompatible with the diagonal form of the actual difference between the functions. 
This rather cumbersome derivation demonstrates that -- even though the general form of scalar invariants discussed in Ref.~\citenum{vill+21nips} does provide a framework to build universal approximators -- the specific form chosen as a practical example is not able to fit certain ground truth functional dependencies for certain types of 3D point clouds. 

\vspace{6cm}\

\vspace{3cm}


\begin{thebibliography}{69}%
\makeatletter
\providecommand \@ifxundefined [1]{%
 \@ifx{#1\undefined}
}%
\providecommand \@ifnum [1]{%
 \ifnum #1\expandafter \@firstoftwo
 \else \expandafter \@secondoftwo
 \fi
}%
\providecommand \@ifx [1]{%
 \ifx #1\expandafter \@firstoftwo
 \else \expandafter \@secondoftwo
 \fi
}%
\providecommand \natexlab [1]{#1}%
\providecommand \enquote  [1]{``#1''}%
\providecommand \bibnamefont  [1]{#1}%
\providecommand \bibfnamefont [1]{#1}%
\providecommand \citenamefont [1]{#1}%
\providecommand \href@noop [0]{\@secondoftwo}%
\providecommand \href [0]{\begingroup \@sanitize@url \@href}%
\providecommand \@href[1]{\@@startlink{#1}\@@href}%
\providecommand \@@href[1]{\endgroup#1\@@endlink}%
\providecommand \@sanitize@url [0]{\catcode `\\12\catcode `\$12\catcode
  `\&12\catcode `\#12\catcode `\^12\catcode `\_12\catcode `\%12\relax}%
\providecommand \@@startlink[1]{}%
\providecommand \@@endlink[0]{}%
\providecommand \url  [0]{\begingroup\@sanitize@url \@url }%
\providecommand \@url [1]{\endgroup\@href {#1}{\urlprefix }}%
\providecommand \urlprefix  [0]{URL }%
\providecommand \Eprint [0]{\href }%
\providecommand \doibase [0]{http://dx.doi.org/}%
\providecommand \selectlanguage [0]{\@gobble}%
\providecommand \bibinfo  [0]{\@secondoftwo}%
\providecommand \bibfield  [0]{\@secondoftwo}%
\providecommand \translation [1]{[#1]}%
\providecommand \BibitemOpen [0]{}%
\providecommand \bibitemStop [0]{}%
\providecommand \bibitemNoStop [0]{.\EOS\space}%
\providecommand \EOS [0]{\spacefactor3000\relax}%
\providecommand \BibitemShut  [1]{\csname bibitem#1\endcsname}%
\let\auto@bib@innerbib\@empty
\bibitem [{\citenamefont {Gumhold}\ \emph {et~al.}(2001)\citenamefont
  {Gumhold}, \citenamefont {Wang}, \citenamefont {MacLeod} \emph
  {et~al.}}]{gumh+01imr}%
  \BibitemOpen
  \bibfield  {author} {\bibinfo {author} {\bibfnamefont {S.}~\bibnamefont
  {Gumhold}}, \bibinfo {author} {\bibfnamefont {X.}~\bibnamefont {Wang}},
  \bibinfo {author} {\bibfnamefont {R.~S.}\ \bibnamefont {MacLeod}},  \emph
  {et~al.},\ }in\ \href@noop {} {\emph {\bibinfo {booktitle} {{{IMR}}}}}\
  (\bibinfo {year} {2001})\ pp.\ \bibinfo {pages} {293--305}\BibitemShut
  {NoStop}%
\bibitem [{\citenamefont {Guo}\ \emph {et~al.}(2021)\citenamefont {Guo},
  \citenamefont {Wang}, \citenamefont {Hu}, \citenamefont {Liu}, \citenamefont
  {Liu},\ and\ \citenamefont {Bennamoun}}]{guo+21tpami}%
  \BibitemOpen
  \bibfield  {author} {\bibinfo {author} {\bibfnamefont {Y.}~\bibnamefont
  {Guo}}, \bibinfo {author} {\bibfnamefont {H.}~\bibnamefont {Wang}}, \bibinfo
  {author} {\bibfnamefont {Q.}~\bibnamefont {Hu}}, \bibinfo {author}
  {\bibfnamefont {H.}~\bibnamefont {Liu}}, \bibinfo {author} {\bibfnamefont
  {L.}~\bibnamefont {Liu}}, \ and\ \bibinfo {author} {\bibfnamefont
  {M.}~\bibnamefont {Bennamoun}},\ }\href {\doibase 10.1109/TPAMI.2020.3005434}
  {\bibfield  {journal} {\bibinfo  {journal} {IEEE Trans. Pattern Anal. Mach.
  Intell.}\ }\textbf {\bibinfo {volume} {43}},\ \bibinfo {pages} {4338}
  (\bibinfo {year} {2021})}\BibitemShut {NoStop}%
\bibitem [{\citenamefont {Wu}\ \emph {et~al.}(2019)\citenamefont {Wu},
  \citenamefont {Qi},\ and\ \citenamefont {Fuxin}}]{wu+19ccvpr}%
  \BibitemOpen
  \bibfield  {author} {\bibinfo {author} {\bibfnamefont {W.}~\bibnamefont
  {Wu}}, \bibinfo {author} {\bibfnamefont {Z.}~\bibnamefont {Qi}}, \ and\
  \bibinfo {author} {\bibfnamefont {L.}~\bibnamefont {Fuxin}},\ }in\ \href
  {\doibase 10.1109/CVPR.2019.00985} {\emph {\bibinfo {booktitle} {2019
  {{IEEE}}/{{CVF Conference}} on {{Computer Vision}} and {{Pattern
  Recognition}} ({{CVPR}})}}}\ (\bibinfo  {publisher} {{IEEE}},\ \bibinfo
  {address} {{Long Beach, CA, USA}},\ \bibinfo {year} {2019})\ pp.\ \bibinfo
  {pages} {9613--9622}\BibitemShut {NoStop}%
\bibitem [{\citenamefont {Bello}\ \emph {et~al.}(2020)\citenamefont {Bello},
  \citenamefont {Yu}, \citenamefont {Wang}, \citenamefont {Adam},\ and\
  \citenamefont {Li}}]{bell+20rs}%
  \BibitemOpen
  \bibfield  {author} {\bibinfo {author} {\bibfnamefont {S.~A.}\ \bibnamefont
  {Bello}}, \bibinfo {author} {\bibfnamefont {S.}~\bibnamefont {Yu}}, \bibinfo
  {author} {\bibfnamefont {C.}~\bibnamefont {Wang}}, \bibinfo {author}
  {\bibfnamefont {J.~M.}\ \bibnamefont {Adam}}, \ and\ \bibinfo {author}
  {\bibfnamefont {J.}~\bibnamefont {Li}},\ }\href {\doibase 10.3390/rs12111729}
  {\bibfield  {journal} {\bibinfo  {journal} {Remote Sensing}\ }\textbf
  {\bibinfo {volume} {12}},\ \bibinfo {pages} {1729} (\bibinfo {year}
  {2020})}\BibitemShut {NoStop}%
\bibitem [{\citenamefont {Li}\ \emph {et~al.}(2021)\citenamefont {Li},
  \citenamefont {Ma}, \citenamefont {Zhong}, \citenamefont {Liu}, \citenamefont
  {Chapman}, \citenamefont {Cao},\ and\ \citenamefont {Li}}]{li+21tnnls}%
  \BibitemOpen
  \bibfield  {author} {\bibinfo {author} {\bibfnamefont {Y.}~\bibnamefont
  {Li}}, \bibinfo {author} {\bibfnamefont {L.}~\bibnamefont {Ma}}, \bibinfo
  {author} {\bibfnamefont {Z.}~\bibnamefont {Zhong}}, \bibinfo {author}
  {\bibfnamefont {F.}~\bibnamefont {Liu}}, \bibinfo {author} {\bibfnamefont
  {M.~A.}\ \bibnamefont {Chapman}}, \bibinfo {author} {\bibfnamefont
  {D.}~\bibnamefont {Cao}}, \ and\ \bibinfo {author} {\bibfnamefont
  {J.}~\bibnamefont {Li}},\ }\href {\doibase 10.1109/TNNLS.2020.3015992}
  {\bibfield  {journal} {\bibinfo  {journal} {IEEE Trans. Neural Netw. Learning
  Syst.}\ }\textbf {\bibinfo {volume} {32}},\ \bibinfo {pages} {3412} (\bibinfo
  {year} {2021})}\BibitemShut {NoStop}%
\bibitem [{\citenamefont {Behler}\ and\ \citenamefont
  {Parrinello}(2007)}]{behl-parr07prl}%
  \BibitemOpen
  \bibfield  {author} {\bibinfo {author} {\bibfnamefont {J.}~\bibnamefont
  {Behler}}\ and\ \bibinfo {author} {\bibfnamefont {M.}~\bibnamefont
  {Parrinello}},\ }\href {\doibase 10.1103/PhysRevLett.98.146401} {\bibfield
  {journal} {\bibinfo  {journal} {Phys. Rev. Lett.}\ }\textbf {\bibinfo
  {volume} {98}},\ \bibinfo {pages} {146401} (\bibinfo {year}
  {2007})}\BibitemShut {NoStop}%
\bibitem [{\citenamefont {Bart{\'o}k}\ \emph {et~al.}(2010)\citenamefont
  {Bart{\'o}k}, \citenamefont {Payne}, \citenamefont {Kondor},\ and\
  \citenamefont {Cs{\'a}nyi}}]{bart+10prl}%
  \BibitemOpen
  \bibfield  {author} {\bibinfo {author} {\bibfnamefont {A.~P.}\ \bibnamefont
  {Bart{\'o}k}}, \bibinfo {author} {\bibfnamefont {M.~C.}\ \bibnamefont
  {Payne}}, \bibinfo {author} {\bibfnamefont {R.}~\bibnamefont {Kondor}}, \
  and\ \bibinfo {author} {\bibfnamefont {G.}~\bibnamefont {Cs{\'a}nyi}},\
  }\href {\doibase 10.1103/PhysRevLett.104.136403} {\bibfield  {journal}
  {\bibinfo  {journal} {Phys. Rev. Lett.}\ }\textbf {\bibinfo {volume} {104}},\
  \bibinfo {pages} {136403} (\bibinfo {year} {2010})}\BibitemShut {NoStop}%
\bibitem [{\citenamefont {Rupp}\ \emph {et~al.}(2012)\citenamefont {Rupp},
  \citenamefont {Tkatchenko}, \citenamefont {M{\"u}ller},\ and\ \citenamefont
  {{von Lilienfeld}}}]{rupp+12prl}%
  \BibitemOpen
  \bibfield  {author} {\bibinfo {author} {\bibfnamefont {M.}~\bibnamefont
  {Rupp}}, \bibinfo {author} {\bibfnamefont {A.}~\bibnamefont {Tkatchenko}},
  \bibinfo {author} {\bibfnamefont {K.-R.}\ \bibnamefont {M{\"u}ller}}, \ and\
  \bibinfo {author} {\bibfnamefont {O.~A.}\ \bibnamefont {{von Lilienfeld}}},\
  }\href {\doibase 10.1103/PhysRevLett.108.058301} {\bibfield  {journal}
  {\bibinfo  {journal} {Phys. Rev. Lett.}\ }\textbf {\bibinfo {volume} {108}},\
  \bibinfo {pages} {058301} (\bibinfo {year} {2012})}\BibitemShut {NoStop}%
\bibitem [{\citenamefont {Musil}\ \emph {et~al.}(2021)\citenamefont {Musil},
  \citenamefont {Grisafi}, \citenamefont {Bart{\'o}k}, \citenamefont {Ortner},
  \citenamefont {Cs{\'a}nyi},\ and\ \citenamefont {Ceriotti}}]{musi+21cr}%
  \BibitemOpen
  \bibfield  {author} {\bibinfo {author} {\bibfnamefont {F.}~\bibnamefont
  {Musil}}, \bibinfo {author} {\bibfnamefont {A.}~\bibnamefont {Grisafi}},
  \bibinfo {author} {\bibfnamefont {A.~P.}\ \bibnamefont {Bart{\'o}k}},
  \bibinfo {author} {\bibfnamefont {C.}~\bibnamefont {Ortner}}, \bibinfo
  {author} {\bibfnamefont {G.}~\bibnamefont {Cs{\'a}nyi}}, \ and\ \bibinfo
  {author} {\bibfnamefont {M.}~\bibnamefont {Ceriotti}},\ }\href {\doibase
  10.1021/acs.chemrev.1c00021} {\bibfield  {journal} {\bibinfo  {journal}
  {Chem. Rev.}\ }\textbf {\bibinfo {volume} {121}},\ \bibinfo {pages} {9759}
  (\bibinfo {year} {2021})}\BibitemShut {NoStop}%
\bibitem [{\citenamefont {Willatt}\ \emph {et~al.}(2019)\citenamefont
  {Willatt}, \citenamefont {Musil},\ and\ \citenamefont
  {Ceriotti}}]{will+19jcp}%
  \BibitemOpen
  \bibfield  {author} {\bibinfo {author} {\bibfnamefont {M.~J.}\ \bibnamefont
  {Willatt}}, \bibinfo {author} {\bibfnamefont {F.}~\bibnamefont {Musil}}, \
  and\ \bibinfo {author} {\bibfnamefont {M.}~\bibnamefont {Ceriotti}},\ }\href
  {\doibase 10.1063/1.5090481} {\bibfield  {journal} {\bibinfo  {journal} {J.
  Chem. Phys.}\ }\textbf {\bibinfo {volume} {150}},\ \bibinfo {pages} {154110}
  (\bibinfo {year} {2019})}\BibitemShut {NoStop}%
\bibitem [{\citenamefont {Pozdnyakov}\ \emph {et~al.}(2020)\citenamefont
  {Pozdnyakov}, \citenamefont {Willatt}, \citenamefont {Bart{\'o}k},
  \citenamefont {Ortner}, \citenamefont {Cs{\'a}nyi},\ and\ \citenamefont
  {Ceriotti}}]{pozd+20prl}%
  \BibitemOpen
  \bibfield  {author} {\bibinfo {author} {\bibfnamefont {S.~N.}\ \bibnamefont
  {Pozdnyakov}}, \bibinfo {author} {\bibfnamefont {M.~J.}\ \bibnamefont
  {Willatt}}, \bibinfo {author} {\bibfnamefont {A.~P.}\ \bibnamefont
  {Bart{\'o}k}}, \bibinfo {author} {\bibfnamefont {C.}~\bibnamefont {Ortner}},
  \bibinfo {author} {\bibfnamefont {G.}~\bibnamefont {Cs{\'a}nyi}}, \ and\
  \bibinfo {author} {\bibfnamefont {M.}~\bibnamefont {Ceriotti}},\ }\href
  {\doibase 10.1103/PhysRevLett.125.166001} {\bibfield  {journal} {\bibinfo
  {journal} {Phys. Rev. Lett.}\ }\textbf {\bibinfo {volume} {125}},\ \bibinfo
  {pages} {166001} (\bibinfo {year} {2020})}\BibitemShut {NoStop}%
\bibitem [{\citenamefont {Pozdnyakov}\ \emph {et~al.}(2021)\citenamefont
  {Pozdnyakov}, \citenamefont {Zhang}, \citenamefont {Ortner}, \citenamefont
  {Cs{\'a}nyi},\ and\ \citenamefont {Ceriotti}}]{pozd+21ore}%
  \BibitemOpen
  \bibfield  {author} {\bibinfo {author} {\bibfnamefont {S.~N.}\ \bibnamefont
  {Pozdnyakov}}, \bibinfo {author} {\bibfnamefont {L.}~\bibnamefont {Zhang}},
  \bibinfo {author} {\bibfnamefont {C.}~\bibnamefont {Ortner}}, \bibinfo
  {author} {\bibfnamefont {G.}~\bibnamefont {Cs{\'a}nyi}}, \ and\ \bibinfo
  {author} {\bibfnamefont {M.}~\bibnamefont {Ceriotti}},\ }\href {\doibase
  10.12688/openreseurope.14156.1} {\bibfield  {journal} {\bibinfo  {journal}
  {Open Res Europe}\ }\textbf {\bibinfo {volume} {1}},\ \bibinfo {pages} {126}
  (\bibinfo {year} {2021})}\BibitemShut {NoStop}%
\bibitem [{\citenamefont {Bart{\'o}k}\ \emph {et~al.}(2013)\citenamefont
  {Bart{\'o}k}, \citenamefont {Kondor},\ and\ \citenamefont
  {Cs{\'a}nyi}}]{bart+13prb}%
  \BibitemOpen
  \bibfield  {author} {\bibinfo {author} {\bibfnamefont {A.~P.}\ \bibnamefont
  {Bart{\'o}k}}, \bibinfo {author} {\bibfnamefont {R.}~\bibnamefont {Kondor}},
  \ and\ \bibinfo {author} {\bibfnamefont {G.}~\bibnamefont {Cs{\'a}nyi}},\
  }\href {\doibase 10.1103/PhysRevB.87.184115} {\bibfield  {journal} {\bibinfo
  {journal} {Phys. Rev. B}\ }\textbf {\bibinfo {volume} {87}},\ \bibinfo
  {pages} {184115} (\bibinfo {year} {2013})}\BibitemShut {NoStop}%
\bibitem [{\citenamefont {Boutin}\ and\ \citenamefont
  {Kemper}(2004)}]{bout-kemp04aam}%
  \BibitemOpen
  \bibfield  {author} {\bibinfo {author} {\bibfnamefont {M.}~\bibnamefont
  {Boutin}}\ and\ \bibinfo {author} {\bibfnamefont {G.}~\bibnamefont
  {Kemper}},\ }\href {\doibase 10.1016/S0196-8858(03)00101-5} {\bibfield
  {journal} {\bibinfo  {journal} {Advances in Applied Mathematics}\ }\textbf
  {\bibinfo {volume} {32}},\ \bibinfo {pages} {709} (\bibinfo {year}
  {2004})}\BibitemShut {NoStop}%
\bibitem [{\citenamefont {Sadeghi}\ \emph {et~al.}(2013)\citenamefont
  {Sadeghi}, \citenamefont {Ghasemi}, \citenamefont {Schaefer}, \citenamefont
  {Mohr}, \citenamefont {Lill},\ and\ \citenamefont {Goedecker}}]{sade+13jcp}%
  \BibitemOpen
  \bibfield  {author} {\bibinfo {author} {\bibfnamefont {A.}~\bibnamefont
  {Sadeghi}}, \bibinfo {author} {\bibfnamefont {S.~A.}\ \bibnamefont
  {Ghasemi}}, \bibinfo {author} {\bibfnamefont {B.}~\bibnamefont {Schaefer}},
  \bibinfo {author} {\bibfnamefont {S.}~\bibnamefont {Mohr}}, \bibinfo {author}
  {\bibfnamefont {M.~A.}\ \bibnamefont {Lill}}, \ and\ \bibinfo {author}
  {\bibfnamefont {S.}~\bibnamefont {Goedecker}},\ }\href {\doibase
  10.1063/1.4828704} {\bibfield  {journal} {\bibinfo  {journal} {J. Chem.
  Phys.}\ }\textbf {\bibinfo {volume} {139}},\ \bibinfo {pages} {184118}
  (\bibinfo {year} {2013})}\BibitemShut {NoStop}%
\bibitem [{\citenamefont {Widdowson}\ \emph {et~al.}(2022)\citenamefont
  {Widdowson}, \citenamefont {Mosca}, \citenamefont {Pulido}, \citenamefont
  {Cooper},\ and\ \citenamefont {Kurlin}}]{widd+22cmcc}%
  \BibitemOpen
  \bibfield  {author} {\bibinfo {author} {\bibfnamefont {D.}~\bibnamefont
  {Widdowson}}, \bibinfo {author} {\bibfnamefont {M.~M.}\ \bibnamefont
  {Mosca}}, \bibinfo {author} {\bibfnamefont {A.}~\bibnamefont {Pulido}},
  \bibinfo {author} {\bibfnamefont {A.~I.}\ \bibnamefont {Cooper}}, \ and\
  \bibinfo {author} {\bibfnamefont {V.}~\bibnamefont {Kurlin}},\ }\href
  {\doibase 10.46793/match.87-3.529W} {\bibfield  {journal} {\bibinfo
  {journal} {match}\ }\textbf {\bibinfo {volume} {87}},\ \bibinfo {pages} {529}
  (\bibinfo {year} {2022})}\BibitemShut {NoStop}%
\bibitem [{\citenamefont {{von Lilienfeld}}\ \emph {et~al.}(2015)\citenamefont
  {{von Lilienfeld}}, \citenamefont {Ramakrishnan}, \citenamefont {Rupp},\ and\
  \citenamefont {Knoll}}]{vonl+15ijqc}%
  \BibitemOpen
  \bibfield  {author} {\bibinfo {author} {\bibfnamefont {O.~A.}\ \bibnamefont
  {{von Lilienfeld}}}, \bibinfo {author} {\bibfnamefont {R.}~\bibnamefont
  {Ramakrishnan}}, \bibinfo {author} {\bibfnamefont {M.}~\bibnamefont {Rupp}},
  \ and\ \bibinfo {author} {\bibfnamefont {A.}~\bibnamefont {Knoll}},\ }\href
  {\doibase 10.1002/qua.24912} {\bibfield  {journal} {\bibinfo  {journal} {Int.
  J. Quantum Chem.}\ }\textbf {\bibinfo {volume} {115}},\ \bibinfo {pages}
  {1084} (\bibinfo {year} {2015})}\BibitemShut {NoStop}%
\bibitem [{\citenamefont {Shapeev}(2016)}]{shap16mms}%
  \BibitemOpen
  \bibfield  {author} {\bibinfo {author} {\bibfnamefont {A.~V.}\ \bibnamefont
  {Shapeev}},\ }\href {\doibase 10.1137/15M1054183} {\bibfield  {journal}
  {\bibinfo  {journal} {Multiscale Model. Simul.}\ }\textbf {\bibinfo {volume}
  {14}},\ \bibinfo {pages} {1153} (\bibinfo {year} {2016})}\BibitemShut
  {NoStop}%
\bibitem [{\citenamefont {Drautz}(2019)}]{drau19prb}%
  \BibitemOpen
  \bibfield  {author} {\bibinfo {author} {\bibfnamefont {R.}~\bibnamefont
  {Drautz}},\ }\href {\doibase 10.1103/PhysRevB.99.014104} {\bibfield
  {journal} {\bibinfo  {journal} {Phys. Rev. B}\ }\textbf {\bibinfo {volume}
  {99}},\ \bibinfo {pages} {014104} (\bibinfo {year} {2019})}\BibitemShut
  {NoStop}%
\bibitem [{\citenamefont {Nigam}\ \emph {et~al.}(2020)\citenamefont {Nigam},
  \citenamefont {Pozdnyakov},\ and\ \citenamefont {Ceriotti}}]{niga+20jcp}%
  \BibitemOpen
  \bibfield  {author} {\bibinfo {author} {\bibfnamefont {J.}~\bibnamefont
  {Nigam}}, \bibinfo {author} {\bibfnamefont {S.}~\bibnamefont {Pozdnyakov}}, \
  and\ \bibinfo {author} {\bibfnamefont {M.}~\bibnamefont {Ceriotti}},\ }\href
  {\doibase 10.1063/5.0021116} {\bibfield  {journal} {\bibinfo  {journal} {J.
  Chem. Phys.}\ }\textbf {\bibinfo {volume} {153}},\ \bibinfo {pages} {121101}
  (\bibinfo {year} {2020})}\BibitemShut {NoStop}%
\bibitem [{\citenamefont {Wang}\ \emph {et~al.}(2019)\citenamefont {Wang},
  \citenamefont {Sun}, \citenamefont {Liu}, \citenamefont {Sarma},
  \citenamefont {Bronstein},\ and\ \citenamefont {Solomon}}]{wang+19acmtg}%
  \BibitemOpen
  \bibfield  {author} {\bibinfo {author} {\bibfnamefont {Y.}~\bibnamefont
  {Wang}}, \bibinfo {author} {\bibfnamefont {Y.}~\bibnamefont {Sun}}, \bibinfo
  {author} {\bibfnamefont {Z.}~\bibnamefont {Liu}}, \bibinfo {author}
  {\bibfnamefont {S.~E.}\ \bibnamefont {Sarma}}, \bibinfo {author}
  {\bibfnamefont {M.~M.}\ \bibnamefont {Bronstein}}, \ and\ \bibinfo {author}
  {\bibfnamefont {J.~M.}\ \bibnamefont {Solomon}},\ }\href {\doibase
  10.1145/3326362} {\bibfield  {journal} {\bibinfo  {journal} {ACM Trans.
  Graph.}\ }\textbf {\bibinfo {volume} {38}},\ \bibinfo {pages} {1} (\bibinfo
  {year} {2019})}\BibitemShut {NoStop}%
\bibitem [{\citenamefont {Gilmer}\ \emph {et~al.}(2017)\citenamefont {Gilmer},
  \citenamefont {Schoenholz}, \citenamefont {Riley}, \citenamefont {Vinyals},\
  and\ \citenamefont {Dahl}}]{gilm+17icml}%
  \BibitemOpen
  \bibfield  {author} {\bibinfo {author} {\bibfnamefont {J.}~\bibnamefont
  {Gilmer}}, \bibinfo {author} {\bibfnamefont {S.~S.}\ \bibnamefont
  {Schoenholz}}, \bibinfo {author} {\bibfnamefont {P.~F.}\ \bibnamefont
  {Riley}}, \bibinfo {author} {\bibfnamefont {O.}~\bibnamefont {Vinyals}}, \
  and\ \bibinfo {author} {\bibfnamefont {G.~E.}\ \bibnamefont {Dahl}},\ }in\
  \href@noop {} {\emph {\bibinfo {booktitle} {International Conference on
  Machine Learning}}}\ (\bibinfo {year} {2017})\ pp.\ \bibinfo {pages}
  {1263--1272}\BibitemShut {NoStop}%
\bibitem [{\citenamefont {Sch{\"u}tt}\ \emph {et~al.}(2018)\citenamefont
  {Sch{\"u}tt}, \citenamefont {Sauceda}, \citenamefont {Kindermans},
  \citenamefont {Tkatchenko},\ and\ \citenamefont {M{\"u}ller}}]{schu+18jcp}%
  \BibitemOpen
  \bibfield  {author} {\bibinfo {author} {\bibfnamefont {K.~T.}\ \bibnamefont
  {Sch{\"u}tt}}, \bibinfo {author} {\bibfnamefont {H.~E.}\ \bibnamefont
  {Sauceda}}, \bibinfo {author} {\bibfnamefont {P.-J.}\ \bibnamefont
  {Kindermans}}, \bibinfo {author} {\bibfnamefont {A.}~\bibnamefont
  {Tkatchenko}}, \ and\ \bibinfo {author} {\bibfnamefont {K.-R.}\ \bibnamefont
  {M{\"u}ller}},\ }\href {\doibase 10.1063/1.5019779} {\bibfield  {journal}
  {\bibinfo  {journal} {J. Chem. Phys.}\ }\textbf {\bibinfo {volume} {148}},\
  \bibinfo {pages} {241722} (\bibinfo {year} {2018})}\BibitemShut {NoStop}%
\bibitem [{\citenamefont {Sch{\"u}tt}\ \emph {et~al.}(2017)\citenamefont
  {Sch{\"u}tt}, \citenamefont {Kindermans}, \citenamefont {Felix},
  \citenamefont {Chmiela}, \citenamefont {Tkatchenko},\ and\ \citenamefont
  {M{\"u}ller}}]{schu+17nips}%
  \BibitemOpen
  \bibfield  {author} {\bibinfo {author} {\bibfnamefont {K.}~\bibnamefont
  {Sch{\"u}tt}}, \bibinfo {author} {\bibfnamefont {P.-J.}\ \bibnamefont
  {Kindermans}}, \bibinfo {author} {\bibfnamefont {H.~E.~S.}\ \bibnamefont
  {Felix}}, \bibinfo {author} {\bibfnamefont {S.}~\bibnamefont {Chmiela}},
  \bibinfo {author} {\bibfnamefont {A.}~\bibnamefont {Tkatchenko}}, \ and\
  \bibinfo {author} {\bibfnamefont {K.-R.}\ \bibnamefont {M{\"u}ller}},\ }in\
  \href {\doibase 10.5555/3294771.3294866} {\emph {\bibinfo {booktitle}
  {{{NIPS}}}}}\ (\bibinfo {year} {2017})\BibitemShut {NoStop}%
\bibitem [{\citenamefont {Sato}(2020)}]{sato2020arxiv}%
  \BibitemOpen
  \bibfield  {author} {\bibinfo {author} {\bibfnamefont {R.}~\bibnamefont
  {Sato}},\ }\href {http://arxiv.org/abs/2003.04078v4} {\bibfield  {journal}
  {\bibinfo  {journal} {arxiv:2003.04078}\ } (\bibinfo {year}
  {2020})}\BibitemShut {NoStop}%
\bibitem [{\citenamefont {Morris}\ \emph {et~al.}(2021)\citenamefont {Morris},
  \citenamefont {Ritzert}, \citenamefont {Fey}, \citenamefont {Hamilton},
  \citenamefont {Lenssen}, \citenamefont {Rattan},\ and\ \citenamefont
  {Grohe}}]{morr+18arxiv}%
  \BibitemOpen
  \bibfield  {author} {\bibinfo {author} {\bibfnamefont {C.}~\bibnamefont
  {Morris}}, \bibinfo {author} {\bibfnamefont {M.}~\bibnamefont {Ritzert}},
  \bibinfo {author} {\bibfnamefont {M.}~\bibnamefont {Fey}}, \bibinfo {author}
  {\bibfnamefont {W.~L.}\ \bibnamefont {Hamilton}}, \bibinfo {author}
  {\bibfnamefont {J.~E.}\ \bibnamefont {Lenssen}}, \bibinfo {author}
  {\bibfnamefont {G.}~\bibnamefont {Rattan}}, \ and\ \bibinfo {author}
  {\bibfnamefont {M.}~\bibnamefont {Grohe}},\ }\href@noop {} {\bibfield
  {journal} {\bibinfo  {journal} {ArXiv181002244 Cs Stat}\ } (\bibinfo {year}
  {2021})}\BibitemShut {NoStop}%
\bibitem [{\citenamefont {Thomas}\ \emph {et~al.}(2018)\citenamefont {Thomas},
  \citenamefont {Smidt}, \citenamefont {Kearnes}, \citenamefont {Yang},
  \citenamefont {Li}, \citenamefont {Kohlhoff},\ and\ \citenamefont
  {Riley}}]{thom+18arxiv}%
  \BibitemOpen
  \bibfield  {author} {\bibinfo {author} {\bibfnamefont {N.}~\bibnamefont
  {Thomas}}, \bibinfo {author} {\bibfnamefont {T.}~\bibnamefont {Smidt}},
  \bibinfo {author} {\bibfnamefont {S.}~\bibnamefont {Kearnes}}, \bibinfo
  {author} {\bibfnamefont {L.}~\bibnamefont {Yang}}, \bibinfo {author}
  {\bibfnamefont {L.}~\bibnamefont {Li}}, \bibinfo {author} {\bibfnamefont
  {K.}~\bibnamefont {Kohlhoff}}, \ and\ \bibinfo {author} {\bibfnamefont
  {P.}~\bibnamefont {Riley}},\ }\href@noop {} {\bibfield  {journal} {\bibinfo
  {journal} {arxiv:1802.08219}\ } (\bibinfo {year} {2018})}\BibitemShut
  {NoStop}%
\bibitem [{\citenamefont {Zhang}\ \emph {et~al.}(2021)\citenamefont {Zhang},
  \citenamefont {Xia},\ and\ \citenamefont {Jiang}}]{zhan+21prl}%
  \BibitemOpen
  \bibfield  {author} {\bibinfo {author} {\bibfnamefont {Y.}~\bibnamefont
  {Zhang}}, \bibinfo {author} {\bibfnamefont {J.}~\bibnamefont {Xia}}, \ and\
  \bibinfo {author} {\bibfnamefont {B.}~\bibnamefont {Jiang}},\ }\href
  {\doibase 10.1103/PhysRevLett.127.156002} {\bibfield  {journal} {\bibinfo
  {journal} {Phys. Rev. Lett.}\ }\textbf {\bibinfo {volume} {127}},\ \bibinfo
  {pages} {156002} (\bibinfo {year} {2021})}\BibitemShut {NoStop}%
\bibitem [{\citenamefont {Garg}\ \emph {et~al.}(2020)\citenamefont {Garg},
  \citenamefont {Jegelka},\ and\ \citenamefont {Jaakkola}}]{vikas2020arxiv}%
  \BibitemOpen
  \bibfield  {author} {\bibinfo {author} {\bibfnamefont {V.~K.}\ \bibnamefont
  {Garg}}, \bibinfo {author} {\bibfnamefont {S.}~\bibnamefont {Jegelka}}, \
  and\ \bibinfo {author} {\bibfnamefont {T.}~\bibnamefont {Jaakkola}},\ }\href
  {http://arxiv.org/abs/2002.06157v1} {\bibfield  {journal} {\bibinfo
  {journal} {arxiv:2002.06157}\ } (\bibinfo {year} {2020})}\BibitemShut
  {NoStop}%
\bibitem [{\citenamefont {Weisfeiler}\ and\ \citenamefont
  {Leman}(1968)}]{weis-lehm68NTI}%
  \BibitemOpen
  \bibfield  {author} {\bibinfo {author} {\bibfnamefont {B.}~\bibnamefont
  {Weisfeiler}}\ and\ \bibinfo {author} {\bibfnamefont {A.}~\bibnamefont
  {Leman}},\ }\href@noop {} {\bibfield  {journal} {\bibinfo  {journal} {NTI
  Ser.}\ }\textbf {\bibinfo {volume} {2}},\ \bibinfo {pages} {12} (\bibinfo
  {year} {1968})}\BibitemShut {NoStop}%
\bibitem [{\citenamefont {Douglas}(2011)}]{douglas2011arxiv}%
  \BibitemOpen
  \bibfield  {author} {\bibinfo {author} {\bibfnamefont {B.~L.}\ \bibnamefont
  {Douglas}},\ }\href {http://arxiv.org/abs/1101.5211v1} {\bibfield  {journal}
  {\bibinfo  {journal} {arxiv:1101.5211}\ } (\bibinfo {year}
  {2011})}\BibitemShut {NoStop}%
\bibitem [{\citenamefont {Shervashidze}\ and\ \citenamefont
  {Borgwardt}(2009)}]{sher-borg09nips}%
  \BibitemOpen
  \bibfield  {author} {\bibinfo {author} {\bibfnamefont {N.}~\bibnamefont
  {Shervashidze}}\ and\ \bibinfo {author} {\bibfnamefont {K.~M.}\ \bibnamefont
  {Borgwardt}},\ }in\ \href@noop {} {\emph {\bibinfo {booktitle} {Proceedings
  of the 22nd International Conference on Neural Information Processing
  Systems}}},\ \bibinfo {series and number} {{{NIPS}}'09}\ (\bibinfo
  {publisher} {{Curran Associates Inc.}},\ \bibinfo {address} {{Red Hook, NY,
  USA}},\ \bibinfo {year} {2009})\ pp.\ \bibinfo {pages}
  {1660--1668}\BibitemShut {NoStop}%
\bibitem [{\citenamefont {Kipf}\ and\ \citenamefont
  {Welling}(2016)}]{kipf2016arxiv}%
  \BibitemOpen
  \bibfield  {author} {\bibinfo {author} {\bibfnamefont {T.~N.}\ \bibnamefont
  {Kipf}}\ and\ \bibinfo {author} {\bibfnamefont {M.}~\bibnamefont {Welling}},\
  }\href@noop {} {\bibfield  {journal} {\bibinfo  {journal} {arXiv preprint
  arXiv:1609.02907}\ } (\bibinfo {year} {2016})}\BibitemShut {NoStop}%
\bibitem [{\citenamefont {Shervashidze}\ \emph {et~al.}(2011)\citenamefont
  {Shervashidze}, \citenamefont {Schweitzer}, \citenamefont {{van Leeuwen}},
  \citenamefont {Mehlhorn},\ and\ \citenamefont {Borgwardt}}]{sher+11jmlr}%
  \BibitemOpen
  \bibfield  {author} {\bibinfo {author} {\bibfnamefont {N.}~\bibnamefont
  {Shervashidze}}, \bibinfo {author} {\bibfnamefont {P.}~\bibnamefont
  {Schweitzer}}, \bibinfo {author} {\bibfnamefont {E.~J.}\ \bibnamefont {{van
  Leeuwen}}}, \bibinfo {author} {\bibfnamefont {K.}~\bibnamefont {Mehlhorn}}, \
  and\ \bibinfo {author} {\bibfnamefont {K.~M.}\ \bibnamefont {Borgwardt}},\
  }\href@noop {} {\bibfield  {journal} {\bibinfo  {journal} {J. Mach. Learn.
  Res.}\ }\textbf {\bibinfo {volume} {12}},\ \bibinfo {pages} {2539} (\bibinfo
  {year} {2011})}\BibitemShut {NoStop}%
\bibitem [{\citenamefont {Duvenaud}\ \emph {et~al.}(2015)\citenamefont
  {Duvenaud}, \citenamefont {Maclaurin}, \citenamefont {Iparraguirre},
  \citenamefont {Bombarell}, \citenamefont {Hirzel}, \citenamefont
  {{Aspuru-Guzik}},\ and\ \citenamefont {Adams}}]{duve+15nips}%
  \BibitemOpen
  \bibfield  {author} {\bibinfo {author} {\bibfnamefont {D.~K.}\ \bibnamefont
  {Duvenaud}}, \bibinfo {author} {\bibfnamefont {D.}~\bibnamefont {Maclaurin}},
  \bibinfo {author} {\bibfnamefont {J.}~\bibnamefont {Iparraguirre}}, \bibinfo
  {author} {\bibfnamefont {R.}~\bibnamefont {Bombarell}}, \bibinfo {author}
  {\bibfnamefont {T.}~\bibnamefont {Hirzel}}, \bibinfo {author} {\bibfnamefont
  {A.}~\bibnamefont {{Aspuru-Guzik}}}, \ and\ \bibinfo {author} {\bibfnamefont
  {R.~P.}\ \bibnamefont {Adams}},\ }in\ \href@noop {} {\emph {\bibinfo
  {booktitle} {Advances in Neural Information Processing Systems}}}\ (\bibinfo
  {year} {2015})\ pp.\ \bibinfo {pages} {2224--2232}\BibitemShut {NoStop}%
\bibitem [{\citenamefont {Kearnes}\ \emph {et~al.}(2016)\citenamefont
  {Kearnes}, \citenamefont {McCloskey}, \citenamefont {Berndl}, \citenamefont
  {Pande},\ and\ \citenamefont {Riley}}]{kear+16jcamd}%
  \BibitemOpen
  \bibfield  {author} {\bibinfo {author} {\bibfnamefont {S.}~\bibnamefont
  {Kearnes}}, \bibinfo {author} {\bibfnamefont {K.}~\bibnamefont {McCloskey}},
  \bibinfo {author} {\bibfnamefont {M.}~\bibnamefont {Berndl}}, \bibinfo
  {author} {\bibfnamefont {V.}~\bibnamefont {Pande}}, \ and\ \bibinfo {author}
  {\bibfnamefont {P.}~\bibnamefont {Riley}},\ }\href {\doibase
  10.1007/s10822-016-9938-8} {\bibfield  {journal} {\bibinfo  {journal} {J
  Comput Aided Mol Des}\ }\textbf {\bibinfo {volume} {30}},\ \bibinfo {pages}
  {595} (\bibinfo {year} {2016})}\BibitemShut {NoStop}%
\bibitem [{\citenamefont {Chen}\ \emph {et~al.}(2019)\citenamefont {Chen},
  \citenamefont {Ye}, \citenamefont {Zuo}, \citenamefont {Zheng},\ and\
  \citenamefont {Ong}}]{chen+19cm}%
  \BibitemOpen
  \bibfield  {author} {\bibinfo {author} {\bibfnamefont {C.}~\bibnamefont
  {Chen}}, \bibinfo {author} {\bibfnamefont {W.}~\bibnamefont {Ye}}, \bibinfo
  {author} {\bibfnamefont {Y.}~\bibnamefont {Zuo}}, \bibinfo {author}
  {\bibfnamefont {C.}~\bibnamefont {Zheng}}, \ and\ \bibinfo {author}
  {\bibfnamefont {S.~P.}\ \bibnamefont {Ong}},\ }\href {\doibase
  10.1021/acs.chemmater.9b01294} {\bibfield  {journal} {\bibinfo  {journal}
  {Chem. Mater.}\ }\textbf {\bibinfo {volume} {31}},\ \bibinfo {pages} {3564}
  (\bibinfo {year} {2019})}\BibitemShut {NoStop}%
\bibitem [{\citenamefont {Xu}\ \emph {et~al.}(2019)\citenamefont {Xu},
  \citenamefont {Hu}, \citenamefont {Leskovec},\ and\ \citenamefont
  {Jegelka}}]{xu2018how}%
  \BibitemOpen
  \bibfield  {author} {\bibinfo {author} {\bibfnamefont {K.}~\bibnamefont
  {Xu}}, \bibinfo {author} {\bibfnamefont {W.}~\bibnamefont {Hu}}, \bibinfo
  {author} {\bibfnamefont {J.}~\bibnamefont {Leskovec}}, \ and\ \bibinfo
  {author} {\bibfnamefont {S.}~\bibnamefont {Jegelka}},\ }in\ \href
  {https://openreview.net/forum?id=ryGs6iA5Km} {\emph {\bibinfo {booktitle}
  {International Conference on Learning Representations}}}\ (\bibinfo {year}
  {2019})\BibitemShut {NoStop}%
\bibitem [{\citenamefont {Parsaeifard}\ and\ \citenamefont
  {Goedecker}(2022)}]{pars-goed22jcp}%
  \BibitemOpen
  \bibfield  {author} {\bibinfo {author} {\bibfnamefont {B.}~\bibnamefont
  {Parsaeifard}}\ and\ \bibinfo {author} {\bibfnamefont {S.}~\bibnamefont
  {Goedecker}},\ }\href {\doibase 10.1063/5.0070488} {\bibfield  {journal}
  {\bibinfo  {journal} {J. Chem. Phys.}\ }\textbf {\bibinfo {volume} {156}},\
  \bibinfo {pages} {034302} (\bibinfo {year} {2022})}\BibitemShut {NoStop}%
\bibitem [{\citenamefont {Pozdnyakov}\ \emph {et~al.}(2022)\citenamefont
  {Pozdnyakov}, \citenamefont {Willatt}, \citenamefont {Bartok-Partay},
  \citenamefont {Ortner}, \citenamefont {Csanyi},\ and\ \citenamefont
  {Ceriotti}}]{pozd+22jcp}%
  \BibitemOpen
  \bibfield  {author} {\bibinfo {author} {\bibfnamefont {S.}~\bibnamefont
  {Pozdnyakov}}, \bibinfo {author} {\bibfnamefont {M.~J.}\ \bibnamefont
  {Willatt}}, \bibinfo {author} {\bibfnamefont {A.}~\bibnamefont
  {Bartok-Partay}}, \bibinfo {author} {\bibfnamefont {C.}~\bibnamefont
  {Ortner}}, \bibinfo {author} {\bibfnamefont {G.}~\bibnamefont {Csanyi}}, \
  and\ \bibinfo {author} {\bibfnamefont {M.}~\bibnamefont {Ceriotti}},\ }\href
  {\doibase 10.1063/5.0088404} {\bibfield  {journal} {\bibinfo  {journal} {J.
  Chem. Phys.}\ } (\bibinfo {year} {2022}),\ 10.1063/5.0088404}\BibitemShut
  {NoStop}%
\bibitem [{\citenamefont {Gasteiger}\ \emph {et~al.}(2020)\citenamefont
  {Gasteiger}, \citenamefont {Groß},\ and\ \citenamefont
  {Günnemann}}]{Gasteiger2020Directional}%
  \BibitemOpen
  \bibfield  {author} {\bibinfo {author} {\bibfnamefont {J.}~\bibnamefont
  {Gasteiger}}, \bibinfo {author} {\bibfnamefont {J.}~\bibnamefont {Groß}}, \
  and\ \bibinfo {author} {\bibfnamefont {S.}~\bibnamefont {Günnemann}},\ }in\
  \href {https://openreview.net/forum?id=B1eWbxStPH} {\emph {\bibinfo
  {booktitle} {International Conference on Learning Representations}}}\
  (\bibinfo {year} {2020})\BibitemShut {NoStop}%
\bibitem [{\citenamefont {Klicpera}\ \emph {et~al.}(2021)\citenamefont
  {Klicpera}, \citenamefont {Becker},\ and\ \citenamefont
  {G{\"u}nnemann}}]{klic+21arxiv}%
  \BibitemOpen
  \bibfield  {author} {\bibinfo {author} {\bibfnamefont {J.}~\bibnamefont
  {Klicpera}}, \bibinfo {author} {\bibfnamefont {F.}~\bibnamefont {Becker}}, \
  and\ \bibinfo {author} {\bibfnamefont {S.}~\bibnamefont {G{\"u}nnemann}},\
  }\href@noop {} {\bibfield  {journal} {\bibinfo  {journal} {arxiv:2106.08903}\
  } (\bibinfo {year} {2021})}\BibitemShut {NoStop}%
\bibitem [{\citenamefont {Zhang}\ \emph {et~al.}(2022)\citenamefont {Zhang},
  \citenamefont {Xia},\ and\ \citenamefont {Jiang}}]{10.1063/5.0080766}%
  \BibitemOpen
  \bibfield  {author} {\bibinfo {author} {\bibfnamefont {Y.}~\bibnamefont
  {Zhang}}, \bibinfo {author} {\bibfnamefont {J.}~\bibnamefont {Xia}}, \ and\
  \bibinfo {author} {\bibfnamefont {B.}~\bibnamefont {Jiang}},\ }\href
  {\doibase 10.1063/5.0080766} {\bibfield  {journal} {\bibinfo  {journal} {The
  Journal of Chemical Physics}\ }\textbf {\bibinfo {volume} {156}},\ \bibinfo
  {pages} {114801} (\bibinfo {year} {2022})},\ \Eprint
  {http://arxiv.org/abs/https://doi.org/10.1063/5.0080766}
  {https://doi.org/10.1063/5.0080766} \BibitemShut {NoStop}%
\bibitem [{\citenamefont {Choudhary}\ and\ \citenamefont
  {DeCost}(2021)}]{chou-deco21npjcm}%
  \BibitemOpen
  \bibfield  {author} {\bibinfo {author} {\bibfnamefont {K.}~\bibnamefont
  {Choudhary}}\ and\ \bibinfo {author} {\bibfnamefont {B.}~\bibnamefont
  {DeCost}},\ }\href {\doibase 10.1038/s41524-021-00650-1} {\bibfield
  {journal} {\bibinfo  {journal} {npj Comput Mater}\ }\textbf {\bibinfo
  {volume} {7}},\ \bibinfo {pages} {185} (\bibinfo {year} {2021})}\BibitemShut
  {NoStop}%
\bibitem [{\citenamefont {Sanchez}\ \emph {et~al.}(1984)\citenamefont
  {Sanchez}, \citenamefont {Ducastelle},\ and\ \citenamefont
  {Gratias}}]{sanc+84pa}%
  \BibitemOpen
  \bibfield  {author} {\bibinfo {author} {\bibfnamefont {J.}~\bibnamefont
  {Sanchez}}, \bibinfo {author} {\bibfnamefont {F.}~\bibnamefont {Ducastelle}},
  \ and\ \bibinfo {author} {\bibfnamefont {D.}~\bibnamefont {Gratias}},\ }\href
  {\doibase 10.1016/0378-4371(84)90096-7} {\bibfield  {journal} {\bibinfo
  {journal} {Physica A: Statistical Mechanics and its Applications}\ }\textbf
  {\bibinfo {volume} {128}},\ \bibinfo {pages} {334} (\bibinfo {year}
  {1984})}\BibitemShut {NoStop}%
\bibitem [{\citenamefont {Dym}\ and\ \citenamefont
  {Maron}(2020)}]{dym-maro20arxiv}%
  \BibitemOpen
  \bibfield  {author} {\bibinfo {author} {\bibfnamefont {N.}~\bibnamefont
  {Dym}}\ and\ \bibinfo {author} {\bibfnamefont {H.}~\bibnamefont {Maron}},\
  }\href@noop {} {\bibfield  {journal} {\bibinfo  {journal} {arxiv:2010.02449}\
  } (\bibinfo {year} {2020})}\BibitemShut {NoStop}%
\bibitem [{\citenamefont {Anderson}\ \emph {et~al.}(2019)\citenamefont
  {Anderson}, \citenamefont {Hy},\ and\ \citenamefont {Kondor}}]{ande+19nips}%
  \BibitemOpen
  \bibfield  {author} {\bibinfo {author} {\bibfnamefont {B.}~\bibnamefont
  {Anderson}}, \bibinfo {author} {\bibfnamefont {T.~S.}\ \bibnamefont {Hy}}, \
  and\ \bibinfo {author} {\bibfnamefont {R.}~\bibnamefont {Kondor}},\ }in\
  \href@noop {} {\emph {\bibinfo {booktitle} {{{NeurIPS}}}}}\ (\bibinfo {year}
  {2019})\ p.~\bibinfo {pages} {10}\BibitemShut {NoStop}%
\bibitem [{\citenamefont {Nigam}\ \emph {et~al.}(2022)\citenamefont {Nigam},
  \citenamefont {Pozdnyakov}, \citenamefont {Fraux},\ and\ \citenamefont
  {Ceriotti}}]{niga+22jcp2}%
  \BibitemOpen
  \bibfield  {author} {\bibinfo {author} {\bibfnamefont {J.}~\bibnamefont
  {Nigam}}, \bibinfo {author} {\bibfnamefont {S.}~\bibnamefont {Pozdnyakov}},
  \bibinfo {author} {\bibfnamefont {G.}~\bibnamefont {Fraux}}, \ and\ \bibinfo
  {author} {\bibfnamefont {M.}~\bibnamefont {Ceriotti}},\ }\href {\doibase
  10.1063/5.0087042} {\bibfield  {journal} {\bibinfo  {journal} {J. Chem.
  Phys.}\ }\textbf {\bibinfo {volume} {156}},\ \bibinfo {pages} {204115}
  (\bibinfo {year} {2022})}\BibitemShut {NoStop}%
\bibitem [{\citenamefont {Villar}\ \emph {et~al.}(2021)\citenamefont {Villar},
  \citenamefont {Hogg}, \citenamefont {Storey-Fisher}, \citenamefont {Yao},\
  and\ \citenamefont {Blum-Smith}}]{vill+21nips}%
  \BibitemOpen
  \bibfield  {author} {\bibinfo {author} {\bibfnamefont {S.}~\bibnamefont
  {Villar}}, \bibinfo {author} {\bibfnamefont {D.~W.}\ \bibnamefont {Hogg}},
  \bibinfo {author} {\bibfnamefont {K.}~\bibnamefont {Storey-Fisher}}, \bibinfo
  {author} {\bibfnamefont {W.}~\bibnamefont {Yao}}, \ and\ \bibinfo {author}
  {\bibfnamefont {B.}~\bibnamefont {Blum-Smith}},\ }in\ \href
  {https://proceedings.neurips.cc/paper/2021/file/f1b0775946bc0329b35b823b86eeb5f5-Paper.pdf}
  {\emph {\bibinfo {booktitle} {Advances in Neural Information Processing
  Systems}}},\ Vol.~\bibinfo {volume} {34},\ \bibinfo {editor} {edited by\
  \bibinfo {editor} {\bibfnamefont {M.}~\bibnamefont {Ranzato}}, \bibinfo
  {editor} {\bibfnamefont {A.}~\bibnamefont {Beygelzimer}}, \bibinfo {editor}
  {\bibfnamefont {Y.}~\bibnamefont {Dauphin}}, \bibinfo {editor} {\bibfnamefont
  {P.}~\bibnamefont {Liang}}, \ and\ \bibinfo {editor} {\bibfnamefont {J.~W.}\
  \bibnamefont {Vaughan}}}\ (\bibinfo  {publisher} {Curran Associates, Inc.},\
  \bibinfo {year} {2021})\ pp.\ \bibinfo {pages} {28848--28863}\BibitemShut
  {NoStop}%
\bibitem [{\citenamefont {Wales}\ and\ \citenamefont
  {Walsh}(1997)}]{wale-wals97jcp}%
  \BibitemOpen
  \bibfield  {author} {\bibinfo {author} {\bibfnamefont {D.~J.}\ \bibnamefont
  {Wales}}\ and\ \bibinfo {author} {\bibfnamefont {T.~R.}\ \bibnamefont
  {Walsh}},\ }\href {\doibase 10.1063/1.473681} {\bibfield  {journal} {\bibinfo
   {journal} {The Journal of Chemical Physics}\ }\textbf {\bibinfo {volume}
  {106}},\ \bibinfo {pages} {7193} (\bibinfo {year} {1997})}\BibitemShut
  {NoStop}%
\bibitem [{\citenamefont {Maheshwary}\ \emph {et~al.}(2001)\citenamefont
  {Maheshwary}, \citenamefont {Patel}, \citenamefont {Sathyamurthy},
  \citenamefont {Kulkarni},\ and\ \citenamefont {Gadre}}]{mahe+01jpca}%
  \BibitemOpen
  \bibfield  {author} {\bibinfo {author} {\bibfnamefont {S.}~\bibnamefont
  {Maheshwary}}, \bibinfo {author} {\bibfnamefont {N.}~\bibnamefont {Patel}},
  \bibinfo {author} {\bibfnamefont {N.}~\bibnamefont {Sathyamurthy}}, \bibinfo
  {author} {\bibfnamefont {A.~D.}\ \bibnamefont {Kulkarni}}, \ and\ \bibinfo
  {author} {\bibfnamefont {S.~R.}\ \bibnamefont {Gadre}},\ }\href {\doibase
  10.1021/jp013141b} {\bibfield  {journal} {\bibinfo  {journal} {J. Phys. Chem.
  A}\ }\textbf {\bibinfo {volume} {105}},\ \bibinfo {pages} {10525} (\bibinfo
  {year} {2001})}\BibitemShut {NoStop}%
\bibitem [{\citenamefont {Becke}(1993)}]{beck93jcp}%
  \BibitemOpen
  \bibfield  {author} {\bibinfo {author} {\bibfnamefont {A.~D.}\ \bibnamefont
  {Becke}},\ }\href@noop {} {\bibfield  {journal} {\bibinfo  {journal} {J.
  Chem. Phys.}\ }\textbf {\bibinfo {volume} {98}},\ \bibinfo {pages} {5648}
  (\bibinfo {year} {1993})}\BibitemShut {NoStop}%
\bibitem [{\citenamefont {Sun}\ \emph {et~al.}(2020)\citenamefont {Sun},
  \citenamefont {Zhang}, \citenamefont {Banerjee}, \citenamefont {Bao},
  \citenamefont {Barbry}, \citenamefont {Blunt}, \citenamefont {Bogdanov},
  \citenamefont {Booth}, \citenamefont {Chen}, \citenamefont {Cui},
  \citenamefont {Eriksen}, \citenamefont {Gao}, \citenamefont {Guo},
  \citenamefont {Hermann}, \citenamefont {Hermes}, \citenamefont {Koh},
  \citenamefont {Koval}, \citenamefont {Lehtola}, \citenamefont {Li},
  \citenamefont {Liu}, \citenamefont {Mardirossian}, \citenamefont {McClain},
  \citenamefont {Motta}, \citenamefont {Mussard}, \citenamefont {Pham},
  \citenamefont {Pulkin}, \citenamefont {Purwanto}, \citenamefont {Robinson},
  \citenamefont {Ronca}, \citenamefont {Sayfutyarova}, \citenamefont
  {Scheurer}, \citenamefont {Schurkus}, \citenamefont {Smith}, \citenamefont
  {Sun}, \citenamefont {Sun}, \citenamefont {Upadhyay}, \citenamefont {Wagner},
  \citenamefont {Wang}, \citenamefont {White}, \citenamefont {Whitfield},
  \citenamefont {Williamson}, \citenamefont {Wouters}, \citenamefont {Yang},
  \citenamefont {Yu}, \citenamefont {Zhu}, \citenamefont {Berkelbach},
  \citenamefont {Sharma}, \citenamefont {Sokolov},\ and\ \citenamefont
  {Chan}}]{sun+20jcp}%
  \BibitemOpen
  \bibfield  {author} {\bibinfo {author} {\bibfnamefont {Q.}~\bibnamefont
  {Sun}}, \bibinfo {author} {\bibfnamefont {X.}~\bibnamefont {Zhang}}, \bibinfo
  {author} {\bibfnamefont {S.}~\bibnamefont {Banerjee}}, \bibinfo {author}
  {\bibfnamefont {P.}~\bibnamefont {Bao}}, \bibinfo {author} {\bibfnamefont
  {M.}~\bibnamefont {Barbry}}, \bibinfo {author} {\bibfnamefont {N.~S.}\
  \bibnamefont {Blunt}}, \bibinfo {author} {\bibfnamefont {N.~A.}\ \bibnamefont
  {Bogdanov}}, \bibinfo {author} {\bibfnamefont {G.~H.}\ \bibnamefont {Booth}},
  \bibinfo {author} {\bibfnamefont {J.}~\bibnamefont {Chen}}, \bibinfo {author}
  {\bibfnamefont {Z.-H.}\ \bibnamefont {Cui}}, \bibinfo {author} {\bibfnamefont
  {J.~J.}\ \bibnamefont {Eriksen}}, \bibinfo {author} {\bibfnamefont
  {Y.}~\bibnamefont {Gao}}, \bibinfo {author} {\bibfnamefont {S.}~\bibnamefont
  {Guo}}, \bibinfo {author} {\bibfnamefont {J.}~\bibnamefont {Hermann}},
  \bibinfo {author} {\bibfnamefont {M.~R.}\ \bibnamefont {Hermes}}, \bibinfo
  {author} {\bibfnamefont {K.}~\bibnamefont {Koh}}, \bibinfo {author}
  {\bibfnamefont {P.}~\bibnamefont {Koval}}, \bibinfo {author} {\bibfnamefont
  {S.}~\bibnamefont {Lehtola}}, \bibinfo {author} {\bibfnamefont
  {Z.}~\bibnamefont {Li}}, \bibinfo {author} {\bibfnamefont {J.}~\bibnamefont
  {Liu}}, \bibinfo {author} {\bibfnamefont {N.}~\bibnamefont {Mardirossian}},
  \bibinfo {author} {\bibfnamefont {J.~D.}\ \bibnamefont {McClain}}, \bibinfo
  {author} {\bibfnamefont {M.}~\bibnamefont {Motta}}, \bibinfo {author}
  {\bibfnamefont {B.}~\bibnamefont {Mussard}}, \bibinfo {author} {\bibfnamefont
  {H.~Q.}\ \bibnamefont {Pham}}, \bibinfo {author} {\bibfnamefont
  {A.}~\bibnamefont {Pulkin}}, \bibinfo {author} {\bibfnamefont
  {W.}~\bibnamefont {Purwanto}}, \bibinfo {author} {\bibfnamefont {P.~J.}\
  \bibnamefont {Robinson}}, \bibinfo {author} {\bibfnamefont {E.}~\bibnamefont
  {Ronca}}, \bibinfo {author} {\bibfnamefont {E.~R.}\ \bibnamefont
  {Sayfutyarova}}, \bibinfo {author} {\bibfnamefont {M.}~\bibnamefont
  {Scheurer}}, \bibinfo {author} {\bibfnamefont {H.~F.}\ \bibnamefont
  {Schurkus}}, \bibinfo {author} {\bibfnamefont {J.~E.~T.}\ \bibnamefont
  {Smith}}, \bibinfo {author} {\bibfnamefont {C.}~\bibnamefont {Sun}}, \bibinfo
  {author} {\bibfnamefont {S.-N.}\ \bibnamefont {Sun}}, \bibinfo {author}
  {\bibfnamefont {S.}~\bibnamefont {Upadhyay}}, \bibinfo {author}
  {\bibfnamefont {L.~K.}\ \bibnamefont {Wagner}}, \bibinfo {author}
  {\bibfnamefont {X.}~\bibnamefont {Wang}}, \bibinfo {author} {\bibfnamefont
  {A.}~\bibnamefont {White}}, \bibinfo {author} {\bibfnamefont {J.~D.}\
  \bibnamefont {Whitfield}}, \bibinfo {author} {\bibfnamefont {M.~J.}\
  \bibnamefont {Williamson}}, \bibinfo {author} {\bibfnamefont
  {S.}~\bibnamefont {Wouters}}, \bibinfo {author} {\bibfnamefont
  {J.}~\bibnamefont {Yang}}, \bibinfo {author} {\bibfnamefont {J.~M.}\
  \bibnamefont {Yu}}, \bibinfo {author} {\bibfnamefont {T.}~\bibnamefont
  {Zhu}}, \bibinfo {author} {\bibfnamefont {T.~C.}\ \bibnamefont {Berkelbach}},
  \bibinfo {author} {\bibfnamefont {S.}~\bibnamefont {Sharma}}, \bibinfo
  {author} {\bibfnamefont {A.~Y.}\ \bibnamefont {Sokolov}}, \ and\ \bibinfo
  {author} {\bibfnamefont {G.~K.-L.}\ \bibnamefont {Chan}},\ }\href {\doibase
  10.1063/5.0006074} {\bibfield  {journal} {\bibinfo  {journal} {J. Chem.
  Phys.}\ }\textbf {\bibinfo {volume} {153}},\ \bibinfo {pages} {024109}
  (\bibinfo {year} {2020})}\BibitemShut {NoStop}%
\bibitem [{\citenamefont {Liu}\ \emph {et~al.}(1996)\citenamefont {Liu},
  \citenamefont {Cruzan},\ and\ \citenamefont {Saykally}}]{liu+96science}%
  \BibitemOpen
  \bibfield  {author} {\bibinfo {author} {\bibfnamefont {K.}~\bibnamefont
  {Liu}}, \bibinfo {author} {\bibfnamefont {J.~D.}\ \bibnamefont {Cruzan}}, \
  and\ \bibinfo {author} {\bibfnamefont {R.~J.}\ \bibnamefont {Saykally}},\
  }\href {\doibase 10.1126/science.271.5251.929} {\bibfield  {journal}
  {\bibinfo  {journal} {Science}\ }\textbf {\bibinfo {volume} {271}},\ \bibinfo
  {pages} {929} (\bibinfo {year} {1996})}\BibitemShut {NoStop}%
\bibitem [{\citenamefont {Zwier}(2004)}]{zwie04science}%
  \BibitemOpen
  \bibfield  {author} {\bibinfo {author} {\bibfnamefont {T.~S.}\ \bibnamefont
  {Zwier}},\ }\href {\doibase 10.1126/science.1098129} {\bibfield  {journal}
  {\bibinfo  {journal} {Science}\ }\textbf {\bibinfo {volume} {304}},\ \bibinfo
  {pages} {1119} (\bibinfo {year} {2004})}\BibitemShut {NoStop}%
\bibitem [{\citenamefont {Medders}\ \emph {et~al.}(2015)\citenamefont
  {Medders}, \citenamefont {G{\"o}tz}, \citenamefont {Morales}, \citenamefont
  {Bajaj},\ and\ \citenamefont {Paesani}}]{medd+15jcp}%
  \BibitemOpen
  \bibfield  {author} {\bibinfo {author} {\bibfnamefont {G.~R.}\ \bibnamefont
  {Medders}}, \bibinfo {author} {\bibfnamefont {A.~W.}\ \bibnamefont
  {G{\"o}tz}}, \bibinfo {author} {\bibfnamefont {M.~A.}\ \bibnamefont
  {Morales}}, \bibinfo {author} {\bibfnamefont {P.}~\bibnamefont {Bajaj}}, \
  and\ \bibinfo {author} {\bibfnamefont {F.}~\bibnamefont {Paesani}},\ }\href
  {\doibase 10.1063/1.4930194} {\bibfield  {journal} {\bibinfo  {journal} {J.
  Chem. Phys.}\ }\textbf {\bibinfo {volume} {143}},\ \bibinfo {pages} {104102}
  (\bibinfo {year} {2015})}\BibitemShut {NoStop}%
\bibitem [{\citenamefont {Nguyen}\ \emph {et~al.}(2018)\citenamefont {Nguyen},
  \citenamefont {Sz{\'e}kely}, \citenamefont {Imbalzano}, \citenamefont
  {Behler}, \citenamefont {Cs{\'a}nyi}, \citenamefont {Ceriotti}, \citenamefont
  {G{\"o}tz},\ and\ \citenamefont {Paesani}}]{nguy+18jcp}%
  \BibitemOpen
  \bibfield  {author} {\bibinfo {author} {\bibfnamefont {T.~T.}\ \bibnamefont
  {Nguyen}}, \bibinfo {author} {\bibfnamefont {E.}~\bibnamefont {Sz{\'e}kely}},
  \bibinfo {author} {\bibfnamefont {G.}~\bibnamefont {Imbalzano}}, \bibinfo
  {author} {\bibfnamefont {J.}~\bibnamefont {Behler}}, \bibinfo {author}
  {\bibfnamefont {G.}~\bibnamefont {Cs{\'a}nyi}}, \bibinfo {author}
  {\bibfnamefont {M.}~\bibnamefont {Ceriotti}}, \bibinfo {author}
  {\bibfnamefont {A.~W.}\ \bibnamefont {G{\"o}tz}}, \ and\ \bibinfo {author}
  {\bibfnamefont {F.}~\bibnamefont {Paesani}},\ }\href {\doibase
  10.1063/1.5024577} {\bibfield  {journal} {\bibinfo  {journal} {J. Chem.
  Phys.}\ }\textbf {\bibinfo {volume} {148}},\ \bibinfo {pages} {241725}
  (\bibinfo {year} {2018})}\BibitemShut {NoStop}%
\bibitem [{\citenamefont {Heindel}\ and\ \citenamefont
  {Xantheas}(2021)}]{hein-xant21jctc}%
  \BibitemOpen
  \bibfield  {author} {\bibinfo {author} {\bibfnamefont {J.~P.}\ \bibnamefont
  {Heindel}}\ and\ \bibinfo {author} {\bibfnamefont {S.~S.}\ \bibnamefont
  {Xantheas}},\ }\href {\doibase 10.1021/acs.jctc.1c00780} {\bibfield
  {journal} {\bibinfo  {journal} {J. Chem. Theory Comput.}\ }\textbf {\bibinfo
  {volume} {17}},\ \bibinfo {pages} {7341} (\bibinfo {year}
  {2021})}\BibitemShut {NoStop}%
\bibitem [{\citenamefont {Imbalzano}\ \emph {et~al.}(2018)\citenamefont
  {Imbalzano}, \citenamefont {Anelli}, \citenamefont {Giofr{\'e}},
  \citenamefont {Klees}, \citenamefont {Behler},\ and\ \citenamefont
  {Ceriotti}}]{imba+18jcp}%
  \BibitemOpen
  \bibfield  {author} {\bibinfo {author} {\bibfnamefont {G.}~\bibnamefont
  {Imbalzano}}, \bibinfo {author} {\bibfnamefont {A.}~\bibnamefont {Anelli}},
  \bibinfo {author} {\bibfnamefont {D.}~\bibnamefont {Giofr{\'e}}}, \bibinfo
  {author} {\bibfnamefont {S.}~\bibnamefont {Klees}}, \bibinfo {author}
  {\bibfnamefont {J.}~\bibnamefont {Behler}}, \ and\ \bibinfo {author}
  {\bibfnamefont {M.}~\bibnamefont {Ceriotti}},\ }\href {\doibase
  10.1063/1.5024611} {\bibfield  {journal} {\bibinfo  {journal} {J. Chem.
  Phys.}\ }\textbf {\bibinfo {volume} {148}},\ \bibinfo {pages} {241730}
  (\bibinfo {year} {2018})}\BibitemShut {NoStop}%
\bibitem [{\citenamefont {Kapil}\ \emph {et~al.}(2019)\citenamefont {Kapil},
  \citenamefont {Rossi}, \citenamefont {Marsalek}, \citenamefont {Petraglia},
  \citenamefont {Litman}, \citenamefont {Spura}, \citenamefont {Cheng},
  \citenamefont {Cuzzocrea}, \citenamefont {Mei{\ss}ner}, \citenamefont
  {Wilkins}, \citenamefont {Helfrecht}, \citenamefont {Juda}, \citenamefont
  {Bienvenue}, \citenamefont {Fang}, \citenamefont {Kessler}, \citenamefont
  {Poltavsky}, \citenamefont {Vandenbrande}, \citenamefont {Wieme},
  \citenamefont {Corminboeuf}, \citenamefont {K{\"u}hne}, \citenamefont
  {Manolopoulos}, \citenamefont {Markland}, \citenamefont {Richardson},
  \citenamefont {Tkatchenko}, \citenamefont {Tribello}, \citenamefont
  {Van~Speybroeck},\ and\ \citenamefont {Ceriotti}}]{kapi+19cpc}%
  \BibitemOpen
  \bibfield  {author} {\bibinfo {author} {\bibfnamefont {V.}~\bibnamefont
  {Kapil}}, \bibinfo {author} {\bibfnamefont {M.}~\bibnamefont {Rossi}},
  \bibinfo {author} {\bibfnamefont {O.}~\bibnamefont {Marsalek}}, \bibinfo
  {author} {\bibfnamefont {R.}~\bibnamefont {Petraglia}}, \bibinfo {author}
  {\bibfnamefont {Y.}~\bibnamefont {Litman}}, \bibinfo {author} {\bibfnamefont
  {T.}~\bibnamefont {Spura}}, \bibinfo {author} {\bibfnamefont
  {B.}~\bibnamefont {Cheng}}, \bibinfo {author} {\bibfnamefont
  {A.}~\bibnamefont {Cuzzocrea}}, \bibinfo {author} {\bibfnamefont {R.~H.}\
  \bibnamefont {Mei{\ss}ner}}, \bibinfo {author} {\bibfnamefont {D.~M.}\
  \bibnamefont {Wilkins}}, \bibinfo {author} {\bibfnamefont {B.~A.}\
  \bibnamefont {Helfrecht}}, \bibinfo {author} {\bibfnamefont {P.}~\bibnamefont
  {Juda}}, \bibinfo {author} {\bibfnamefont {S.~P.}\ \bibnamefont {Bienvenue}},
  \bibinfo {author} {\bibfnamefont {W.}~\bibnamefont {Fang}}, \bibinfo {author}
  {\bibfnamefont {J.}~\bibnamefont {Kessler}}, \bibinfo {author} {\bibfnamefont
  {I.}~\bibnamefont {Poltavsky}}, \bibinfo {author} {\bibfnamefont
  {S.}~\bibnamefont {Vandenbrande}}, \bibinfo {author} {\bibfnamefont
  {J.}~\bibnamefont {Wieme}}, \bibinfo {author} {\bibfnamefont
  {C.}~\bibnamefont {Corminboeuf}}, \bibinfo {author} {\bibfnamefont {T.~D.}\
  \bibnamefont {K{\"u}hne}}, \bibinfo {author} {\bibfnamefont {D.~E.}\
  \bibnamefont {Manolopoulos}}, \bibinfo {author} {\bibfnamefont {T.~E.}\
  \bibnamefont {Markland}}, \bibinfo {author} {\bibfnamefont {J.~O.}\
  \bibnamefont {Richardson}}, \bibinfo {author} {\bibfnamefont
  {A.}~\bibnamefont {Tkatchenko}}, \bibinfo {author} {\bibfnamefont {G.~A.}\
  \bibnamefont {Tribello}}, \bibinfo {author} {\bibfnamefont {V.}~\bibnamefont
  {Van~Speybroeck}}, \ and\ \bibinfo {author} {\bibfnamefont {M.}~\bibnamefont
  {Ceriotti}},\ }\href {\doibase 10.1016/j.cpc.2018.09.020} {\bibfield
  {journal} {\bibinfo  {journal} {Comput. Phys. Commun.}\ }\textbf {\bibinfo
  {volume} {236}},\ \bibinfo {pages} {214} (\bibinfo {year}
  {2019})}\BibitemShut {NoStop}%
\bibitem [{\citenamefont {Habershon}\ \emph {et~al.}(2009)\citenamefont
  {Habershon}, \citenamefont {Markland},\ and\ \citenamefont
  {Manolopoulos}}]{habe+09jcp}%
  \BibitemOpen
  \bibfield  {author} {\bibinfo {author} {\bibfnamefont {S.}~\bibnamefont
  {Habershon}}, \bibinfo {author} {\bibfnamefont {T.~E.}\ \bibnamefont
  {Markland}}, \ and\ \bibinfo {author} {\bibfnamefont {D.~E.}\ \bibnamefont
  {Manolopoulos}},\ }\href {\doibase 10.1063/1.3167790} {\bibfield  {journal}
  {\bibinfo  {journal} {J. Chem. Phys.}\ }\textbf {\bibinfo {volume} {131}},\
  \bibinfo {pages} {24501} (\bibinfo {year} {2009})}\BibitemShut {NoStop}%
\bibitem [{\citenamefont {Sch{\"u}tt}\ \emph {et~al.}(2019)\citenamefont
  {Sch{\"u}tt}, \citenamefont {Gastegger}, \citenamefont {Tkatchenko},
  \citenamefont {M{\"u}ller},\ and\ \citenamefont {Maurer}}]{schu+19nc}%
  \BibitemOpen
  \bibfield  {author} {\bibinfo {author} {\bibfnamefont {K.~T.}\ \bibnamefont
  {Sch{\"u}tt}}, \bibinfo {author} {\bibfnamefont {M.}~\bibnamefont
  {Gastegger}}, \bibinfo {author} {\bibfnamefont {A.}~\bibnamefont
  {Tkatchenko}}, \bibinfo {author} {\bibfnamefont {K.-R.}\ \bibnamefont
  {M{\"u}ller}}, \ and\ \bibinfo {author} {\bibfnamefont {R.~J.}\ \bibnamefont
  {Maurer}},\ }\href {\doibase 10.1038/s41467-019-12875-2} {\bibfield
  {journal} {\bibinfo  {journal} {Nat Commun}\ }\textbf {\bibinfo {volume}
  {10}},\ \bibinfo {pages} {5024} (\bibinfo {year} {2019})}\BibitemShut
  {NoStop}%
\bibitem [{\citenamefont {Brorsen}(2019)}]{bror19jcp}%
  \BibitemOpen
  \bibfield  {author} {\bibinfo {author} {\bibfnamefont {K.~R.}\ \bibnamefont
  {Brorsen}},\ }\href {\doibase 10.1063/1.5093908} {\bibfield  {journal}
  {\bibinfo  {journal} {J. Chem. Phys.}\ }\textbf {\bibinfo {volume} {150}},\
  \bibinfo {pages} {204104} (\bibinfo {year} {2019})}\BibitemShut {NoStop}%
\bibitem [{\citenamefont {Dandu}\ \emph {et~al.}(2020)\citenamefont {Dandu},
  \citenamefont {Ward}, \citenamefont {Assary}, \citenamefont {Redfern},
  \citenamefont {Narayanan}, \citenamefont {Foster},\ and\ \citenamefont
  {Curtiss}}]{dand+20jpca}%
  \BibitemOpen
  \bibfield  {author} {\bibinfo {author} {\bibfnamefont {N.}~\bibnamefont
  {Dandu}}, \bibinfo {author} {\bibfnamefont {L.}~\bibnamefont {Ward}},
  \bibinfo {author} {\bibfnamefont {R.~S.}\ \bibnamefont {Assary}}, \bibinfo
  {author} {\bibfnamefont {P.~C.}\ \bibnamefont {Redfern}}, \bibinfo {author}
  {\bibfnamefont {B.}~\bibnamefont {Narayanan}}, \bibinfo {author}
  {\bibfnamefont {I.~T.}\ \bibnamefont {Foster}}, \ and\ \bibinfo {author}
  {\bibfnamefont {L.~A.}\ \bibnamefont {Curtiss}},\ }\href {\doibase
  10.1021/acs.jpca.0c01777} {\bibfield  {journal} {\bibinfo  {journal} {J.
  Phys. Chem. A}\ }\textbf {\bibinfo {volume} {124}},\ \bibinfo {pages} {5804}
  (\bibinfo {year} {2020})}\BibitemShut {NoStop}%
\bibitem [{\citenamefont {Westermayr}\ \emph {et~al.}(2020)\citenamefont
  {Westermayr}, \citenamefont {Gastegger},\ and\ \citenamefont
  {Marquetand}}]{west+20jpcl}%
  \BibitemOpen
  \bibfield  {author} {\bibinfo {author} {\bibfnamefont {J.}~\bibnamefont
  {Westermayr}}, \bibinfo {author} {\bibfnamefont {M.}~\bibnamefont
  {Gastegger}}, \ and\ \bibinfo {author} {\bibfnamefont {P.}~\bibnamefont
  {Marquetand}},\ }\href {\doibase 10.1021/acs.jpclett.0c00527} {\bibfield
  {journal} {\bibinfo  {journal} {J. Phys. Chem. Lett.}\ }\textbf {\bibinfo
  {volume} {11}},\ \bibinfo {pages} {3828} (\bibinfo {year}
  {2020})}\BibitemShut {NoStop}%
\bibitem [{\citenamefont {Westermayr}\ and\ \citenamefont
  {Maurer}(2021)}]{west-maur21cs}%
  \BibitemOpen
  \bibfield  {author} {\bibinfo {author} {\bibfnamefont {J.}~\bibnamefont
  {Westermayr}}\ and\ \bibinfo {author} {\bibfnamefont {R.~J.}\ \bibnamefont
  {Maurer}},\ }\href {\doibase 10.1039/D1SC01542G} {\bibfield  {journal}
  {\bibinfo  {journal} {Chem. Sci.}\ }\textbf {\bibinfo {volume} {12}},\
  \bibinfo {pages} {10755} (\bibinfo {year} {2021})}\BibitemShut {NoStop}%
\bibitem [{\citenamefont {Sch{\"u}tt}\ \emph {et~al.}(2021)\citenamefont
  {Sch{\"u}tt}, \citenamefont {Unke},\ and\ \citenamefont
  {Gastegger}}]{pmlr-v139-schutt21a}%
  \BibitemOpen
  \bibfield  {author} {\bibinfo {author} {\bibfnamefont {K.}~\bibnamefont
  {Sch{\"u}tt}}, \bibinfo {author} {\bibfnamefont {O.}~\bibnamefont {Unke}}, \
  and\ \bibinfo {author} {\bibfnamefont {M.}~\bibnamefont {Gastegger}},\ }in\
  \href {https://proceedings.mlr.press/v139/schutt21a.html} {\emph {\bibinfo
  {booktitle} {Proceedings of the 38th International Conference on Machine
  Learning}}},\ \bibinfo {series} {Proceedings of Machine Learning Research},
  Vol.\ \bibinfo {volume} {139},\ \bibinfo {editor} {edited by\ \bibinfo
  {editor} {\bibfnamefont {M.}~\bibnamefont {Meila}}\ and\ \bibinfo {editor}
  {\bibfnamefont {T.}~\bibnamefont {Zhang}}}\ (\bibinfo  {publisher} {PMLR},\
  \bibinfo {year} {2021})\ pp.\ \bibinfo {pages} {9377--9388}\BibitemShut
  {NoStop}%
\bibitem [{\citenamefont {Behler}(2011)}]{behl11jcp}%
  \BibitemOpen
  \bibfield  {author} {\bibinfo {author} {\bibfnamefont {J.}~\bibnamefont
  {Behler}},\ }\href {\doibase 10.1063/1.3553717} {\bibfield  {journal}
  {\bibinfo  {journal} {The Journal of Chemical Physics}\ }\textbf {\bibinfo
  {volume} {134}},\ \bibinfo {pages} {074106} (\bibinfo {year}
  {2011})}\BibitemShut {NoStop}%
\bibitem [{\citenamefont {Kakarala}(2012)}]{kaka12jmiv}%
  \BibitemOpen
  \bibfield  {author} {\bibinfo {author} {\bibfnamefont {R.}~\bibnamefont
  {Kakarala}},\ }\href {\doibase 10.1007/s10851-012-0330-6} {\bibfield
  {journal} {\bibinfo  {journal} {J Math Imaging Vis}\ }\textbf {\bibinfo
  {volume} {44}},\ \bibinfo {pages} {341} (\bibinfo {year} {2012})}\BibitemShut
  {NoStop}%
\end{thebibliography}
\end{document}